\documentclass{article}

\usepackage{arxiv}

\usepackage{graphicx}
\usepackage{hyperref}
\usepackage{amssymb}
\usepackage{bm}

\usepackage{lineno}

\usepackage{amsmath}
\usepackage{diagbox}
\usepackage[ruled,linesnumbered]{algorithm2e}
\SetKwComment{Comment}{$\triangleright$\ }{}
\usepackage{multirow}
\usepackage{array,color}
\usepackage{float}
\usepackage{subcaption}
\usepackage{geometry}

\title{Machine learning based digital twin for stochastic nonlinear multi-degree of freedom dynamical system}

\author{
  Shailesh Garg  \\
  Department of Civil Engineering\\
  Indian Institute of Technology Guwahati\\
  Guwahati, Assam 781039, India. \\
  \texttt{shailesh.garg@iitg.ac.in} \\
   \And
  Ankush Gogoi  \\
  Department of Civil Engineering\\
  Indian Institute of Technology Guwahati\\
  Guwahati, Assam 781039, India. \\
  \texttt{ankushgogoi1@gmail.com} \\
 \And
  Souvik Chakraborty  \\
  Department of Applied Mechanics\\
  Indian Institute of Technology Delhi\\
  Hauz Khas, New Delhi 110016, India. \\
  \texttt{souvik@am.iitd.ac.in} \\
 \And
  Budhaditya Hazra  \\
  Department of Civil Engineering\\
  Indian Institute of Technology Guwahati\\
  Guwahati, Assam 781039, India. \\
  \texttt{budhaditya.hazra@iitg.ac.in} \\
}

\begin{document}
\maketitle

\begin{abstract}
The potential of digital twin technology is immense, specifically in the infrastructure, aerospace, and automotive sector.
However, practical implementation of this technology is not at an expected speed, specifically because of lack of application-specific details.
In this paper, we propose a novel digital twin framework for stochastic nonlinear multi-degree of freedom (MDOF) dynamical systems.
The approach proposed in this paper strategically decouples the problem into two time-scales -- (a) a fast time-scale governing the system dynamics and (b) a slow time-scale governing the degradation in the system.
The proposed digital twin has four components - (a) a physics-based nominal model (low-fidelity), (b) a Bayesian filtering algorithm a (c) a supervised machine learning algorithm
and (d) a high-fidelity model for predicting future responses.
The physics-based nominal model combined with Bayesian filtering is used for combined parameter state estimation and the supervised machine learning algorithm is used for learning the temporal evolution of the parameters.
While the proposed framework can be used with any choice of Bayesian filtering and machine learning algorithm, we propose to use unscented Kalman filter and Gaussian process. 
Performance of the proposed approach is illustrated using two examples. Results obtained indicate the applicability and excellent performance of the proposed digital twin framework.
\end{abstract}


\section{Introduction}
A digital twin (DT) is a digital/virtual representation of a physical system, often referred to as the physical twin.
The virtual model resides in the cloud and is connected to the physical counterpart through internet of things (IoT) \cite{souza2019digital};
the key premise here is to achieve temporal synchronization between the 
physical and digital twins.
This necessitates continuous updation of the DT based on sensor data.
It may be noted that the DT can also actuate the physical counter part through actuators.
Once a DT is in sync with the physical counterpart, it can be used to do a number of tasks including decision making \cite{stassopoulou1998application}, remaining useful-life estimation \cite{si2011remaining} and preventive maintenance optimization \cite{tan1997general}.
The possibilities offered by the DT technology are immense, evident from the recent applications of this technology in prognostics and health monitoring \cite{wang2019digital,booyse2020deep}, manufacturing \cite{lu2020digital,debroy2017building}, automotive and aerospace engineering \cite{li2017dynamic,kapteyn2020toward}, to mention a few.
In this paper, the primary objective is the development of a DT for nonlinear dynamical systems.

Developing DTs for dynamical systems is challenging because of the presence of, at least, two different time-scales.
The responses of dynamical systems are governed by external excitations and their fluctuations thereof; and typically encode the characteristic time period of a system.
On the other hand, the operational life of a dynamical system is rather large. To put things into perspective, for example, consider a wind turbine, where the characteristic time period is of the order of tens (10s) of seconds \cite{adhikari2012dynamic} while the operational life is in tens of years.
Incorporating this fundamental mismatch within a DT is non-trivial.
In \cite{ganguli2020digital}, a digital twin framework for dynamical systems that decouples the two timescales was presented.
Simple analytical formulae for capturing variation of mass and stiffness with operational time-period were proposed.
The framework was later extended in \cite{chakraborty2021role} wherein, a Gaussian Process (GP) \cite{williams1995gaussian,nayek2019gaussian,chakraborty2019graph} was employed for tracking the time evolution of the system parameters.
However, both these frameworks are only applicable when the underlying system is of linear single-degree-of-freedom type.
In practice, most real-life dynamical systems are nonlinear with multiple degrees-of-freedom \cite{tripura2020ito,bhowmik2019first}.
DT for multi-timescale dynamical systems \cite{chakraborty2021machine} and bars \cite{ritto2021digital} can also be found in literature.

From the discussion above, it is evident that the literature on DT for dynamical systems is quite sparse. In fact, to the best of the knowledge of the authors, there exists no work on digital twin for nonlinear MDOF dynamical systems.
In order to fill this apparent void, a novel algorithm for building DT of stochastic nonlinear MDOF dynamical systems is proposed. 
Following \cite{ganguli2020digital}, the proposed framework also decouples the fast and the slow-timescales.
We propose to use nonlinear Bayesian filters \cite{sarkka2013bayesian,chen2003bayesian,yuen2011bayesian} and machine learning \cite{goswami2020transfer,chakraborty2021transfer,chakraborty2016modelling,chakraborty2017polynomial,bilionis2012multi,bilionis2013multi} for estimating the system parameters in the fast and slow timescales, respectively.
It may be noted that the use of Bayesian filters is already quite prevalent for state and parameter estimation.
Works carried out in \cite{nayek2019gaussian,dertimanis2019input,ching2006bayesian,brewick2016probabilistic,saha2009extended} illustrate use of Bayesian filters for state and parameter estimation of linear/non-linear MDOF systems subjected to deterministic loading; however, such filters are only effective over a short time-scale (a few seconds).
The objective here is to develop a DT that can operate over the operational life of a system and hence, directly usage of a Bayesian filter is not an option.
This motivates the coupling Bayesian filter with an appropriate machine learning algorithm.
Among different Bayesian filtering algorithms present in literature, we propose to use unscented Kalman Filter (UKF) \cite{wan2000unscented}.
On the other hand, among different machine learning algorithms existing in the literature, Gaussian process (GP) \cite{nayek2019gaussian,chakraborty2019graph,bilionis2012multi,bilionis2013multi} is used as the machine learning algorithm. The advantage of GP resides in the fact that it is a probabilistic machine learning algorithm and hence, immune from overfitting.
Additionally, it also provides a confidence interval, which is useful in the decision making process. However, one must note that the proposed approach is generic in nature and can be used with any choice of Bayesian filtering and ML algorithms.

The rest of the paper is organized as follows. The dynamic model for the digital twin of nonlinear dynamical systems is discussed in Section \ref{dm}. The problem statement is also stated clearly in this section.
UKF and GP are briefly discussed in Sections \ref{bf} and \ref{gpr} respectively.
The proposed algorithm along with a clear flow-chart is discussed in Section \ref{sec:pa}. Results illustrating the performance of the proposed approach is provided in Section \ref{ni}. Finally, Section \ref{conclusion} provides the concluding remarks.

\section{Dynamic model of the digital twin}\label{dm}
In this section, we present the nominal dynamic system and the DT corresponding to this model.
The nominal model is the `initial model' of a DT.
For structural engineering, we can consider a nominal model to be a numerical model of the system when it is manufactured. 
A DT encapsulates the journey from the nominal model to its updates based on the data collected from the system.
In this section, the key ideas for developing DT of nonlinear MDOF systems are explained.
\subsection{Stochastic nonlinear MDOF system: the nominal model}\label{subsec:nominal_model}
Consider an $N-$DOF stochastic nonlinear system having governing equations as follows:
\begin{equation}\label{eq:nominal}
    \textbf M_0 \ddot {\bm X}  + \mathbf C_0 \dot {\bm X}  + \mathbf K_0 \bm X + \bm G \left( \bm X, \bm \alpha \right) = \bm F + \bm  \Sigma {\bm {\dot{W}}}, 
\end{equation}
where $\mathbf M_0 \in \mathbb R^{N\times N}$, $\mathbf C_0 \in \mathbb R^{N\times N}$ and $\mathbf K_0 \in \mathbb R^{N\times N}$, respectively, represent the  mass, damping and (linear) stiffness matrix of the system.
$\bm G \left( \cdot , \cdot \right) \in \mathbb R^{N}$, on the other hand, represents the nonlinearity present in the system.
$\bm F$ in Eq. (\ref{eq:nominal}) represents the deterministic force and  and $ {\bm {\dot{W}}}$ (Wiener derivative) is the stochastic load vector with noise intensity matrix $\bm \Sigma$.
$\bm \alpha$ in Eq. (\ref{eq:nominal}) represents the parameters corresponding to the nonlinear stiffness model.
Note that $\mathbf M_0$, $\mathbf C_0$ and $\mathbf K_0$ are the nominal parameters and represents the pristine system.
\subsection{The digital twin}
The DT for the N-DOF nonlinear system discussed above can be represented as:
\begin{equation}\label{eq:dt}
    \textbf M(t_s) \frac{\partial^2 \bm X (t,t_s)}{\partial t^2}  + \mathbf C(t_s) \frac{\partial \bm X (t,t_s)}{\partial t}  + \mathbf K (t_s) \bm X (t,t_s) + \bm G \left( \bm (t,t_s), \bm \alpha \right) = \bm F(t, t_s) + \bm \Sigma {\bm {\dot{W}}}, 
\end{equation}
where $t$ represents the system's time and $t_s$ is the service time (operational time-scale).
Note that the response vector $\bm X$ is function of both the time-scales and hence, partial derivatives have been used in Eq. (\ref{eq:dt}).
Eq. (\ref{eq:dt}) is considered to be the DT for the nominal system in Section \ref{subsec:nominal_model}.
Eq. (\ref{eq:dt}) has two time-scales, $t$ and $t_s$.
For all practical purposes the service time-scale $t_s$ is much slower (in months) as compared to the time-scale of the system dynamics.
\subsection{Problem statement}\label{subsec:ps}
Although a physics-based DT for MDOF nonlinear system is defined in Eq. (\ref{eq:dt}), for using it in practice, one needs to estimate the system parameters $\mathbf M(t_s)$, $\mathbf C(t_s)$ and $\mathbf K(t_s)$.
For estimating these parameters, the connectivity between the physical twin and the DT is the key.
Recent developments in IoT provides several new technologies that ensure the connectivity between the two twins.
To be specific, the two-way connectivity between the DT and its counterpart is created by using sensors and actuators.
Given the huge difference in the two times-scales in Eq. (\ref{eq:dt}), it is reasonable to assume that the temporal variation in $\mathbf M(t_s)$, $\mathbf C(t_s)$ and $\mathbf K(t_s)$ are so slow that the dynamics is practically decoupled from these parametric variations.
The sensor collects data intermittently at discrete time instants $t_s$.
At each time instant $t_s$, time history measurements of acceleration response in $t_s \pm \Delta t$ is available.
For this study, it is assumed that there is no practical variation in the mass matrix and hence, $\mathbf M(t_s) = \mathbf M_0$. Variation in damping matrix is also not considered.
With this setup, the objective is to develop a DT for nonlinear MDOF system.
It is envisioned that the DT should be able to track the variation in the system parameters, $\mathbf K(t_s)$ at current time $t$ and is also able to predict future degradation/variation in system parameters.
Last but not the least, a DT should be continuously updated as and when it receives data.

\section{Bayesian Filters}\label{bf}
One of the key components in development of DT is estimating $\mathbf K(t_s)$ given the observations until time $t_s$.
This is a classical parameter estimation problem and this work proposes the use of Bayesian filter to accomplish the goal.
However, one must note that development of DT and parameter estimation are not same; instead, parameter estimation is only a component of the the overall DT.

\begin{figure}[ht!]
    \centering
    \includegraphics[width=0.9\textwidth]{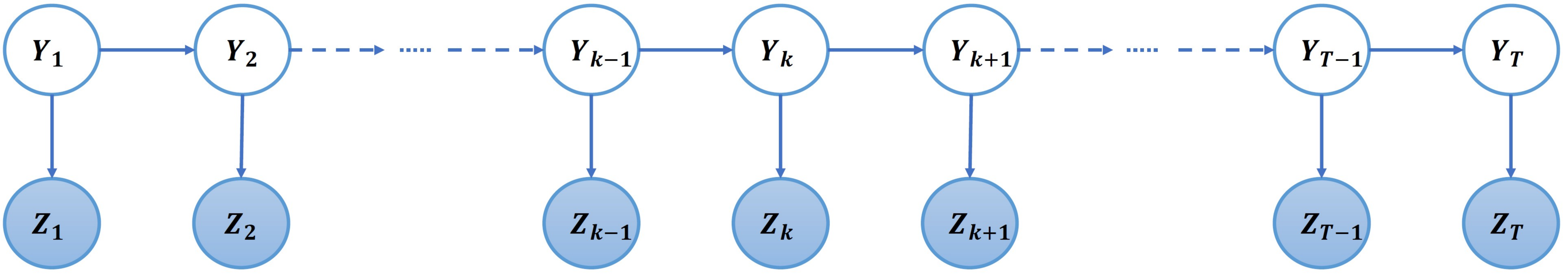}
    \caption{Probabilistic graphical model for state space model. We have considered first order Markovian assumption for the hidden variable $\bm Y$; this ensures that $\bm Y_t$ is only dependent on $\bm Y_{t-1}$.}
    \label{fig:pgm}
\end{figure}

Bayesian filters use Bayesian inference to develop a framework which can then be used for state-parameter estimation. Bayesian inference differs from conventional frequentist approach of statistical inference because it takes probability of an event as the uncertainty of the event in a single trial, as opposed to the proportion of the event in a probability space.
For filtering equations, let the unknown vector be given as $\bm Y_{0:T}=\{\bm Y_0,\bm Y_1,\ldots,\bm Y_T\}$ which is observed through a set of noisy measurements $\bm Z_{1:T}=\{\bm Z_1,\bm Z_2,\ldots, \bm Z_T\}$. Using Bayes's rule,
\begin{equation}
\label{BAYES}
p(\bm Y_{0:T} | \bm Z_{1:T})={\displaystyle\frac{p(\bm Z_{1:T} | \bm Y_{0:T})p(\bm Y_{0:T})}{p(\bm Z_{1:T})}}.
\end{equation}
This full posterior formulation although accurate is computationally heavy and is often intractable.
The computational complexity is simplified by using the first order Markovian assumption.
First order Markov model assumes (i) the state of system at time step $k$ (i.e. $\bm Y_k$) given the state at time step $k-1$ (i.e. $\bm Y_{k-1}$) is independent of anything that has happened before time step $k-1$ and (ii) The measurement at time step $k$ (i.e. $\bm Z_k$) given the state at time step $k$ (i.e. $\bm Y_k$) is independent of any measurement or state histories. Mathematically, this is represented as:
\begin{equation}
p(\bm Y_k | \bm Y_{1:k-1}, \bm Z_{1:k-1})=p(\bm Y_k | \bm Y_{k-1}),
\label{mp1}
\end{equation}
\begin{equation*}
    \text{and}
\end{equation*}
\vspace{-0.45cm}
\begin{equation}
p(\bm Z_k | \bm Y_{1:k},\bm Z_{1:k-1})=p(\bm Z_k | \bm Y_k).
\label{mp2}
\end{equation}
A probabilistic graphical model representing the first-order Markov assumption is shown in Fig. \ref{fig:pgm}. In literature, this is also known as the state-space model (if the state is continuous) or the hidden Markov model (if the state is discrete).
Using assumptions of Markovian model, the recursive Bayesian filter can be set up, and Kalman Filter arises \cite{sarkka2013bayesian, welch1995introduction}, which is a special case of recursive Bayesian filter used for linear models. Extended Kalman Filter \cite{sarkka2013bayesian}, Unscented Kalman Filter (UKF) \cite{sarkka2013bayesian,wan2000unscented} are improvements over Kalman filter, which are used for non-linear models. 
In this work, UKF is used as the Bayesian filtering algorithm of choice.
It is to be noted that UKF is computationally expensive as compared to the EKF algorithm; however, the performance of UKF for systems having higher order of non-linearity is superior \cite{wan2000unscented}.
\subsection{Unscented Kalman Filter}
\begin{figure}[ht!]
    \centering
    \includegraphics[width=0.8\textwidth]{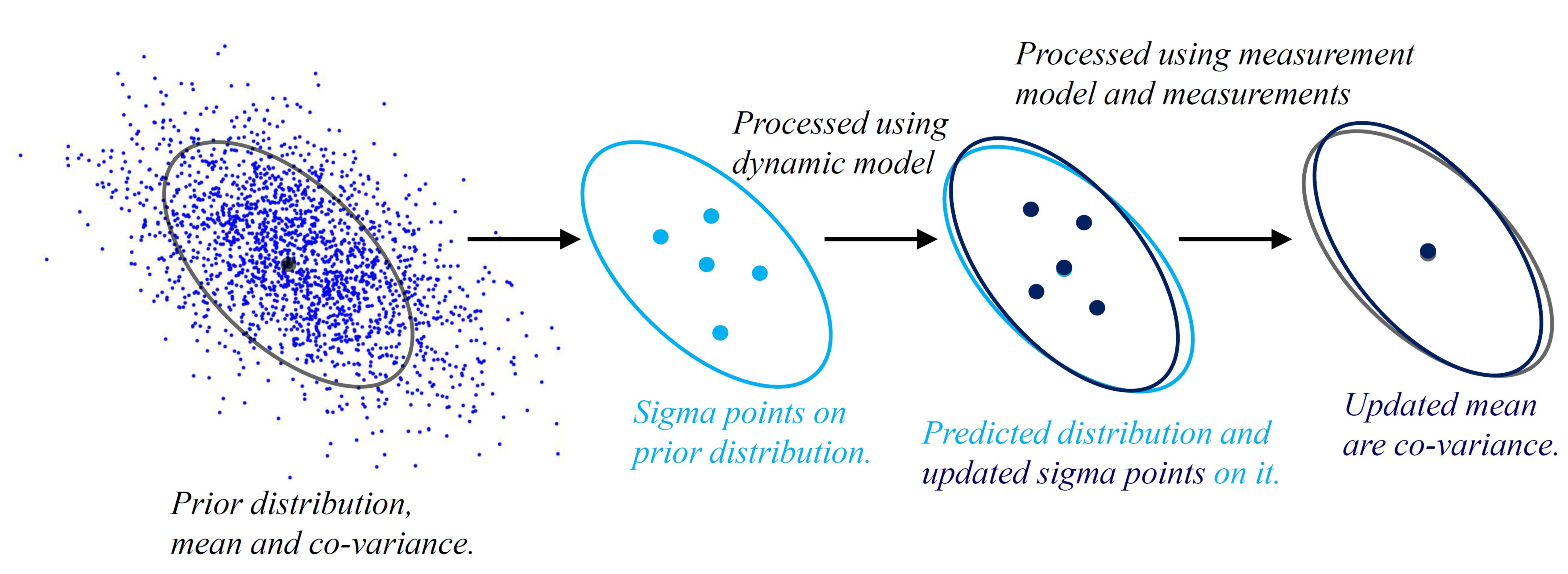}
    \caption{Schematic representation of functionality of sigma points within the UKF framework.}
    \label{fig:sigma}
\end{figure}

UKF uses concepts of unscented transform to analyze non-linear models and directly tries to approximate the mean and co-variance of the targeted distribution instead of approximating the non-linear function. 
To that end, weighted sigma points are used.
The idea here is to consider some points on the source Gaussian distribution which are then mapped onto the target Gaussian distribution after passing through non-linear function.
These points are refereed to as the sigma points and are considered to be representative of the transformed Gaussian distribution.
Considering $L$ to be the length of the state vector, $2L + 1$ sigma points are selected as \cite{wan2000unscented},
\begin{equation}\label{eq:sigma}
\begin{aligned}
    \mathcal{Y}^{(0)} &= \bm \mu\\
    \mathcal{Y}^{(i)} &= \bm \mu+\sqrt{(L+\lambda)}\left[\sqrt{\bm \Sigma}\right],\quad\quad i = 1,.....,L\\
    \mathcal{Y}^{(i)} &= \bm \mu-\sqrt{(L+\lambda)}\left[\sqrt{\bm \Sigma}\right]\quad\quad i = L+1,.....,2L,
\end{aligned}
\end{equation}
where, $\mathcal{Y}$ are the required sigma points, $\bm \mu$ and $\mathbf \Sigma$ are, respectively the mean vector and co-variance matrix. 
$\lambda$ and $L$ in Eq. (\ref{eq:sigma}) represent the scaling parameter and length of state vector respectively.
Details on how to compute $\lambda$ is explained while discussing the UKF algorithm.
Once the mean $m_k$ and covariance $p_k$, are computed using the UKF, we approximate
the filtering distribution as:
\begin{equation}
    p(y_k|z_{1:k})\simeq N(y_k|m_k,p_k),
\end{equation}
where $m_k$ and $p_k$ are the mean and co-variance computed by the algorithm discussed next.
A schematic representation of how sigma points are used within the UKF algorithm is shown in Fig. \ref{fig:sigma}
\subsubsection{Algorithm}\label{sec_ukfa}
\noindent
\textbf{Step 1} Weights calculation for sigma points\par
\quad Select UKF parameters : $\alpha_f = 0.001$, $\beta = 2$, $\kappa = 0$
\begin{equation}
\begin{aligned}
&W_m^{(i = 0)}={\frac{\lambda}{L+\lambda}}\\
&W_c^{(i = 0)}={\frac{\lambda}{L+\lambda}}+(1-\alpha_f^2+\beta),&&i = 1,.....,2L\\
&W_m^{(i)}={\frac{1}{2(L+\lambda)}}\\
&W_c^{(i)}=W_m^{(i)},&&i = 1,.....,2L,
\end{aligned}
\label{ukfw}
\end{equation}
\quad\quad\,\,\,where, $L$ is the length of state vector and scaling parameter $\lambda = \alpha_f^2(L+\kappa)-L$.\\
\textbf{Step 2 : }For k = 0\par
Initialize mean and co-variance i.e. $m_k = m_0$,\,\,$p_k = p_0$.\\
\textbf{Step 3 : }For k = 1,2,.....,$t_n$\par
\textbf{Step 3.1 : }Prediction\par
\quad Getting Sigma points $\mathcal{Y}^{(i)}, i = 0,.....,2L$
\begin{equation}
\begin{aligned}
\mathcal{Y}_{k-1}^{(0)}&=m_{k-1}\\
\mathcal{Y}_{k-1}^{(i)}&=m_{k-1}+\sqrt{L+\lambda}\,\left[\sqrt{P_{k-1}}\right]\\
\mathcal{Y}_{k-1}^{(i+L)}&=m_{k-1}-\sqrt{L+\lambda}\,\left[\sqrt{P_{k-1}}\right],\,\,\,\,\,\,\,\,\,\, i = 1,.....,L.
\end{aligned}
\end{equation}
\quad\quad\,\,\,Propagate sigma points through the dynamic model
\begin{equation}
\mathcal{Y}_k^{(i)}=f(\mathcal{Y}_{k-1}^{(i)}),\,\,\,\,\,\,\,\,\,\,i = 0,.....,2L.
\end{equation}
\quad\quad\,\,\,The predicted mean $m_k^-$ and co-variance $P_k^-$ is then given by
\begin{equation}
\begin{aligned}
m_k^-&=\sum_{i=0}^{2L}W_m^{(i)}\mathcal{Y}_k^{(i)},\\
P_k^-&=\sum_{i=0}^{2L}W_c^{(i)}(\mathcal{Y}_k^{(i)}-m_k^-)(\mathcal{Y}_k^{(i)}-m_k^-)^T+Q_{k-1}.
\end{aligned}
\label{pm-ukf}
\end{equation}
\quad\,\,\,\textbf{Step 3.2 : }Update\par
\quad Getting Sigma points
\begin{equation}
\begin{aligned}
\mathcal{Y}_{k}^{-(0)}&=m_k^-\\
\mathcal{Y}_{k}^{-(i)}&=m_k^-+\sqrt{L+\lambda}\,\left[\sqrt{P_k^-}\right]\\
\mathcal{Y}_{k}^{-(i+L)}&=m_k^--\sqrt{L+\lambda}\,\left[\sqrt{P_k^-}\right],\,\,\,\,\,\,\,\,\,\, i = 1,.....,L.
\end{aligned}
\end{equation}
\quad\quad\,\,\,Propagating sigma points through the measurement model
\begin{equation}
\mathcal{Z}_k^{(i)}=h(\mathcal{Y}_k^{-(i)}),\,\,\,\,\,\,\,\,\,\,i = 0,.....,2L.
\end{equation}
\quad\quad\,\,\,Getting mean $\mu_k$, predicted co-variance $S_k$ and cross co-variance $C_k$
\begin{equation}
\begin{aligned}
\mu_k^-&=\sum_{i=0}^{2L}W_m^{(i)}\mathcal{Z}_k^{(i)},\\
S_k^-&=\sum_{i=0}^{2L}W_c^{(i)}(\mathcal{Z}_k^{(i)}-\mu_k)(\mathcal{Z}_k^{(i)}-\mu_k)^T+R_k,\\
C_k^-&=\sum_{i=0}^{2L}W_c^{(i)}(\mathcal{Y}_k^{-(i)}-m_k^-)(\mathcal{Z}_k^{(i)}-\mu_k)^T.
\end{aligned}
\end{equation}
\quad\,\,\,\textbf{Step 3.3 : }Getting filter gain $K_k$, filtered state mean $m_k$ and co-variance $P_k$ \par
\quad\quad\quad\quad\quad\, conditional on measurement $y_k$.
\begin{equation}
\begin{aligned}
K_k&=C_kS_k^{-1},\\
m_k&=m_k^-+K_k[y_k-\mu_k],\\
P_k&=P_k^--K_kS_kK_k^T.
\end{aligned}
\end{equation}
Within the DT framework, the UKF algorithm is used for parameter estimation at a given time-step, $t_k$.

\section{Gaussian Process Regression}\label{gpr}
In this section, we briefly discuss the other component of the proposed DT framework, namely Gaussian process regression (GPR).
GPR \cite{nayek2019gaussian,bilionis2012multi}, along with neural network \cite{kumar2021state,chakraborty2020simulation} are perhaps the most popular machine learning techniques in today's time.
Unlike conventional frequentist machine learning techniques, GPR doesn't assume a functional form to represent input-output mapping; instead, a distribution over a function is assumed in GPR.
Consequently, GPR has the inherent capability of capturing the epistemic uncertainty \cite{hullermeier2021aleatoric} arising due to limited data.
This feature of GPR is particularly useful when it comes to decision making.
Within the proposed DT framework, we use GPR to track the temporal evolution of the system parameters.

We consider $\bm v_k$ to be the systems parameters and time $\tau_k$.
In GPR, we represent $\bm v_k$ as
\begin{equation}\label{eq:gpr}
    \bm v_k \sim \mathcal G \mathcal P \left( \bm \mu (\tau_k; \bm \beta), \bm \kappa (\tau_k, \tau_k'; \sigma^2, \bm l) \right),
\end{equation}
where $\bm \mu (\cdot; \bm \beta)$ and $\bm \kappa (\cdot, \cdot; \sigma^2, \bm l)$, 
respectively represent the mean function and the covariance function of the GPR.
The mean function is parameterized by the unknown coefficient vector $\bm \beta$ and the covariance function is parameterized by the process variance $\sigma^2$ and the length-scale parameters $\bm l$. All the parameters combined, $\bm \theta = \left[ \bm \beta, \bm l, \sigma ^2 \right]$ are known as hyperparamters of GPR.
It is worthwhile to note that choice of $\bm \mu (\cdot; \bm \beta)$ and $\bm \kappa (\cdot, \cdot; \sigma^2, \bm l)$
has significant influence on the performance of GP; this naturally allows an user to encode prior knowledge into the GPR model and model complex functions \cite{nayek2019gaussian}.
In case there is no prior knowledge about the mean function, it is a common practice to use zero mean Gaussian process,
\begin{equation}\label{eq:zero_gpr}
    \bm v_k \sim \mathcal G \mathcal P \left( \bm 0, \bm \kappa (\tau_k, \tau_k'; \sigma^2, \bm l) \right).
\end{equation}
The covariance function $\bf \kappa (\cdot, \cdot; \sigma^2, \bm l)$, on the other hand, should result in a positive, semi-definite matrix.
For using the GPR in practice, one needs to compute the hyperparameters $\bm \theta$ based on training samples $\mathcal D = \left[ \tau_k, \bm v_k \right]_{k=1}^{N_s}$ where $N_s$ is the number of training samples.
The most widely used method in this regards is based on the maximum likelihood estimation where the negative log-likelihood of GPR is minimized.
For details on MLE for GPR, interested readers may refer \cite{rasmussen2003gaussian}.
The other alternative is to compute the posterior distribution of hyperparameter vector $\bm \theta$ \cite{bilionis2012multi,bilionis2013multi}. This although a superior alternative, renders the process computationally expensive.
In this work, we have used the MLE based approach because of this simplicity.
For ease of readers, the steps involved in training a GPR model are shown in Algorithm \ref{alg:gpr_train}.
\begin{algorithm}[ht!]
\caption{Training GPR}\label{alg:gpr_train}
\textbf{Pre-requisite: }Form of mean function $\bm \mu(\cdot; \bm \beta)$ and covariance function, $\bf \kappa (\cdot, \cdot; \sigma^2, \bm l)$. Provide training data, $\mathcal D = \left[ \tau_k, \bm v_k \right]_{k=1}^{N_s}$, initial values of the parameters, $\bm \theta_0$, maximum allowable iteration $n_{max}$ and error threshold $\epsilon_t$.  \\
$\bm \theta \leftarrow \theta_0$; $iter \leftarrow 0$; $\epsilon \leftarrow 10 \epsilon_t$ \\
\Repeat{$iter \ge n_{max}$ and $\epsilon > \epsilon_t$}{
$iter \leftarrow iter + 1$ \\ 
$\bm \theta _ {iter - 1} \leftarrow  \bm \theta$. \\
Compute the negative log-likelihood by using the training data $\mathcal D$ and $\bm \theta$ \[ f_{ML} \propto \frac{1}{N}\left| \mathbf K \left( \bm \theta \right) + \log \left( \bm v^T \mathbf R \left( \bm \theta \right)^{-1} \bm v \right) \right|,\] where $\mathbf K \left( \bm \theta \right)$ is the covariance matrix computed by using the training data and covariance function $\bf \kappa (\cdot, \cdot; \sigma^2, \bm l)$. $\bm v$ represents the observation vector. \\
Update hyperparameter $\bm \theta$ based on the gradient information. \\
$\bm \theta_{iter} \leftarrow \bm \theta$. \\
$\epsilon \leftarrow \left\| \bm \theta_{iter} - \bm \theta_{iter - 1} \right\|_2^2$
}
\textbf{Output: }Optimal hyper-parameter, $\bm \theta^*$
\end{algorithm}

Once the hyper-parameters $\bm \theta$ are computed, predictive mean and predictive variance corresponding to new input $\tau^*$ are computed as
\begin{equation}\label{eq:pred_mean}
    \bm \mu^* = \bm \Phi \bm \beta^* + \bm \kappa^*(\tau^*;(\sigma^*)2, \bm l^*)\mathbf K^{-1}\left( \bm v - \bm \Phi \bm \beta^* \right),
\end{equation}
\begin{equation}\label{eq:pred_sd}
    s^2(\tau^*) = (\sigma^*)^2\left\{ 1 - \bm \kappa^*(\tau^*;(\sigma^*)2, \bm \theta^*) \mathbf K^{-1} \bm \kappa^*(\tau^*;(\sigma^*)2, \bm l^*)^T + \frac{\left[ 1 - \bm \Phi^T \mathbf K^{-1} \bm \kappa^*(\tau^*;(\sigma^*)2, \bm l^*)^T  \right]}{\bm \Phi^T \mathbf K^{-1} \bm \Phi} \right\},
\end{equation}
where $\bm \beta^*$, $\bm l^*$ and $\sigma^*$ represents the optimized hyper-parameters. $\bm \Phi$ in Eqs. (\ref{eq:pred_mean}) and (\ref{eq:pred_sd}) represents the design matrix.
$\bm \kappa^*(\tau^*;(\sigma^*)2, \bm l^*)^T$ in Eqs. (\ref{eq:pred_mean}) and (\ref{eq:pred_sd}) are the covariance vector between the input training samples and $\tau^*$ and computed as
\begin{equation}
    \bm \kappa^*(\tau^*;(\sigma^*)2, \bm l^*)^T = \left[ \kappa(\tau^*, \tau_1; ;(\sigma^*)2, \bm \theta^*), \ldots, \kappa(\tau^*, \tau_{N_s}; ;(\sigma^*)2, \bm \theta^*) \right].
\end{equation}

\section{Proposed approach}\label{sec:pa}
\begin{figure}[ht!]
    \centering
    \includegraphics[width=0.9\textwidth]{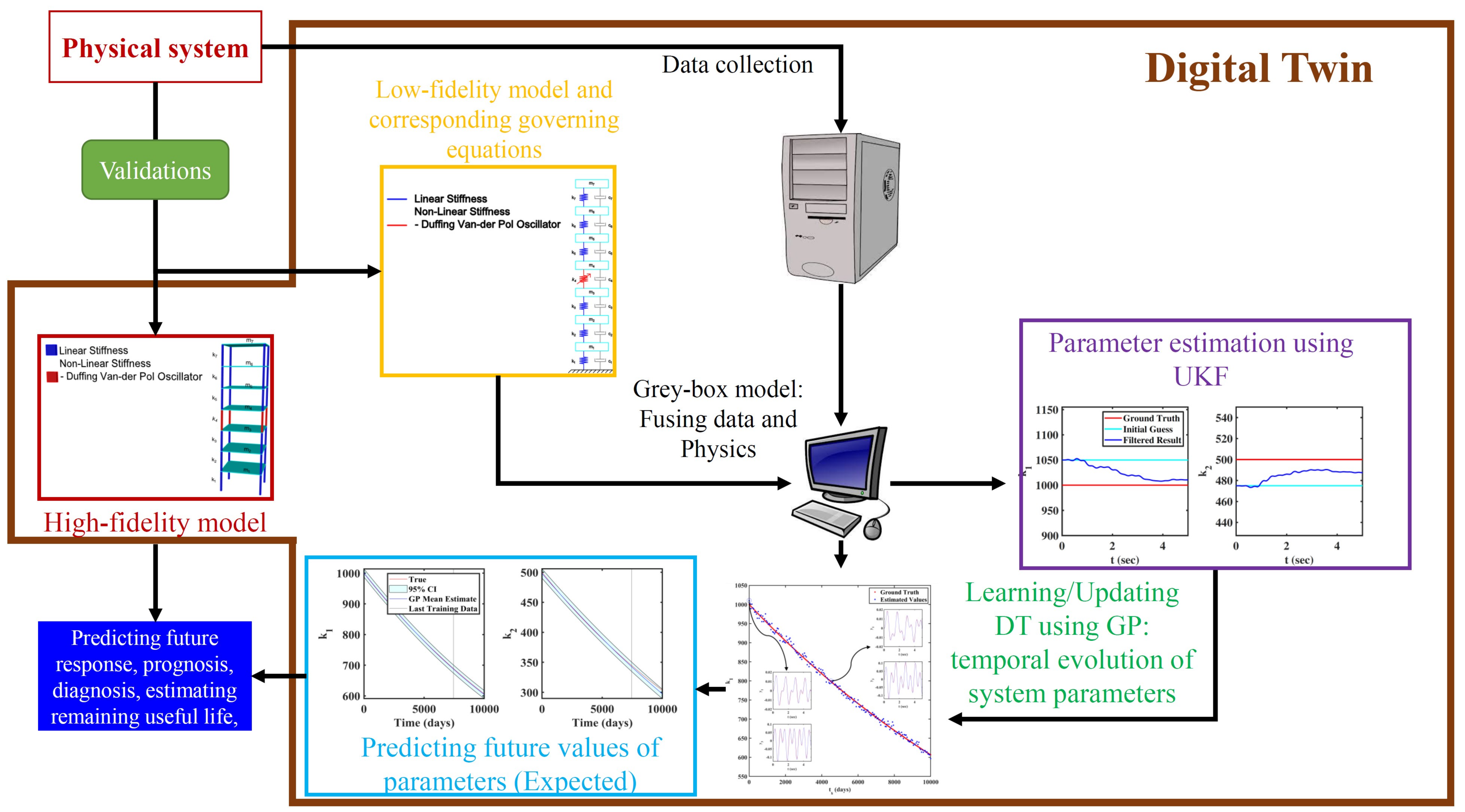}
    \caption{Schematic representation of the proposed digital twin framework. It comprises of low-fidelity model as nominal model, UKF for parameter estimation, GP for learning temporal evolution of parameters and predicting future values of system parameters, and high-fidelity model for estimating future responses.}
    \label{fig:dt}
\end{figure}

Having discussed UKF and GP, the two ingredients of the proposed approach, we proceed to discussing the proposed DT framework for nonlinear dynamical systems.
A schematic representation of the proposed DT is shown in Fig. \ref{fig:dt}.
It has four primary components, namely (a) selection of nominal model, (b) data collection, (c) parameter estimation at a given time-instant and (d) estimation of the temporal variation in parameters.
The selection of nominal model has already been detailed in Section \ref{dm} and hence, here the discussion is limited to data collection, parameter estimation and estimation of temporal variation of the parameters only.

One major concern in DT is its connectivity with the physical counterpart; 
in absence of which, a DT will be of no practical use.
To ensure connectivity, sensors are placed on the physical system (physical twin)
for data collection. The data is communicated to the DT by using cloud technology.
With the substantial advancements in IoT, the access to different types of sensors is straightforward for collecting different types of data.
In this work, we have considered that accelerometers are mounted on the physical system and the DT receives acceleration measurements.
To be specific, one can consider that the acceleration time-history are available to the
DT intermittently at discrete time-instant $t_s$.
Note that the proposed approach is equally applicable (with trivial modifications) if instead of acceleration, displacement or velocity measurements are available. The framework can also be extended to function in tandem with vision based sensors. However, from a practical and economic point-of-view, it is easiest to collect acceleration measurements and hence, the same has been considered in this study.

Once the data is collected, the next objective is to estimate the system parameters (stiffness matrix to be specific), assuming that at time-instant $t_s$, acceleration measurements are avilable in $\left[t_s - \Delta t, t_s  \right]$, where $\Delta t$ is time interval over which acceleration measurement is available at $t_s$.
It is to be noted that $t_s$ is a time-step in the slow time-scale whereas $\Delta t$ is time interval in the fast time-scale.
With this setup, the parameter estimation objective is to estimate  $\mathbf K(t_s)$.
In this work, we estimate $\mathbf K(t_s)$ by using the UKF.
Details on how parameter estimation is carried out using UKF is elaborated in Section \ref{bf}.

The last step within the proposed DT framework is to estimate the temporal evolution of the parameters. 
This is extremely important as it enables the DT to predict future behavior of the 
physical system.
In this work, we propose to use a combination of GPR and UKF for learning the temporal evolution of the system parameters.
To be specific, consider $\bm t = \left[t_1, t_2,\ldots, t^N \right]$ to be time-instants in slow-scale. Also, assume that using UKF, the estimated the system parameters are available at different time-instants as $\bm v = \left[ \bm v_1, \bm v_2, \ldots, \bm v_N \right]$, where $\bm v_i$ includes the elements of stiffness matrix.
The proposed work trains a GPR model between $\bm t$ and $\bm v$,
\begin{equation}\label{eq:gp_dt}
    \bm v \sim \mathcal G \mathcal P (\bm \mu, \bm \kappa).
\end{equation}
Note that for brevity, the hyperparameters in Eq. (\ref{eq:gp_dt}) are omitted.
The GPR is trained by following the procedure discussed ion Algorithm \ref{alg:gpr_train}.
Once trained, the GPR can predict the system parameters at future time-steps.
Note that GPR being a Bayesian machine learning model also provides the predictive uncertainty which can be used to judge the accuracy of the model.
For the ease of readers, the overall DT framework proposed is shown in Algorithm \ref{alg:dt}.
\begin{algorithm}[ht!]
\caption{Proposed DT}\label{alg:dt}
Select nominal model \Comment*[r]{Section \ref{dm}}.
Use data (acceleration measurements) $\mathcal D_s$ collected at time $t_s$ to compute the parameters $\mathbf K(t_s)$ \Comment*[r]{Section \ref{bf}}.
Train a GP using $\mathcal D = \left[ t_n, \bm v_n\right]_{n=1}^{t_s}$ as training data, where $\bm v_n$ represents the system parameter \Comment*[r]{Algorithm \ref{alg:gpr_train}}.
Predict $\mathbf K(\tilde t)$ at future time $\tilde t$\Comment*[r]{Section \ref{gpr}}
Substitute $\mathbf K(\tilde t)$ into the governing equation (high-fidelity model) and solve it to obtain responses at time $\tilde t$. \\
Take decisions related to maintenance, remaining useful life and health of the system. \\
Repeat steps $2-6$ as more data become available
\end{algorithm}

\section{Numerical Illustrations}\label{ni}
In this section, we present two examples to illustrate the performance of the proposed DT framework. 
The first example selected is a 2-DOF system with duffing oscillator attached at the first floor.
As the second example, a 7-DOF system is considered. For this example, the nonlinearity in the system arises because of a duffing van der pol oscillator attached between the third and the fouth DOF.
As stated  earlier, we have considered that acceleration measurements at different time-steps are available.
The objective here is to use the proposed DT to compute the time-evolution of the system parameters.
Once the time-evolution of the parameters are known, the proposed DT can be used for predicting the responses of the system at future time-steps ($t_s$) (see Algorithm \ref{alg:dt} for details).
In this section, we have illustrated how the proposed approach can be used for 
predicting the time-evolution of the system parameters in the past as well as in the future.
\subsection{2-DOF system with duffing oscillator}\label{subsec:duffing}
\begin{figure}[ht!]
    \centering
    \includegraphics[width = 0.4\textwidth]{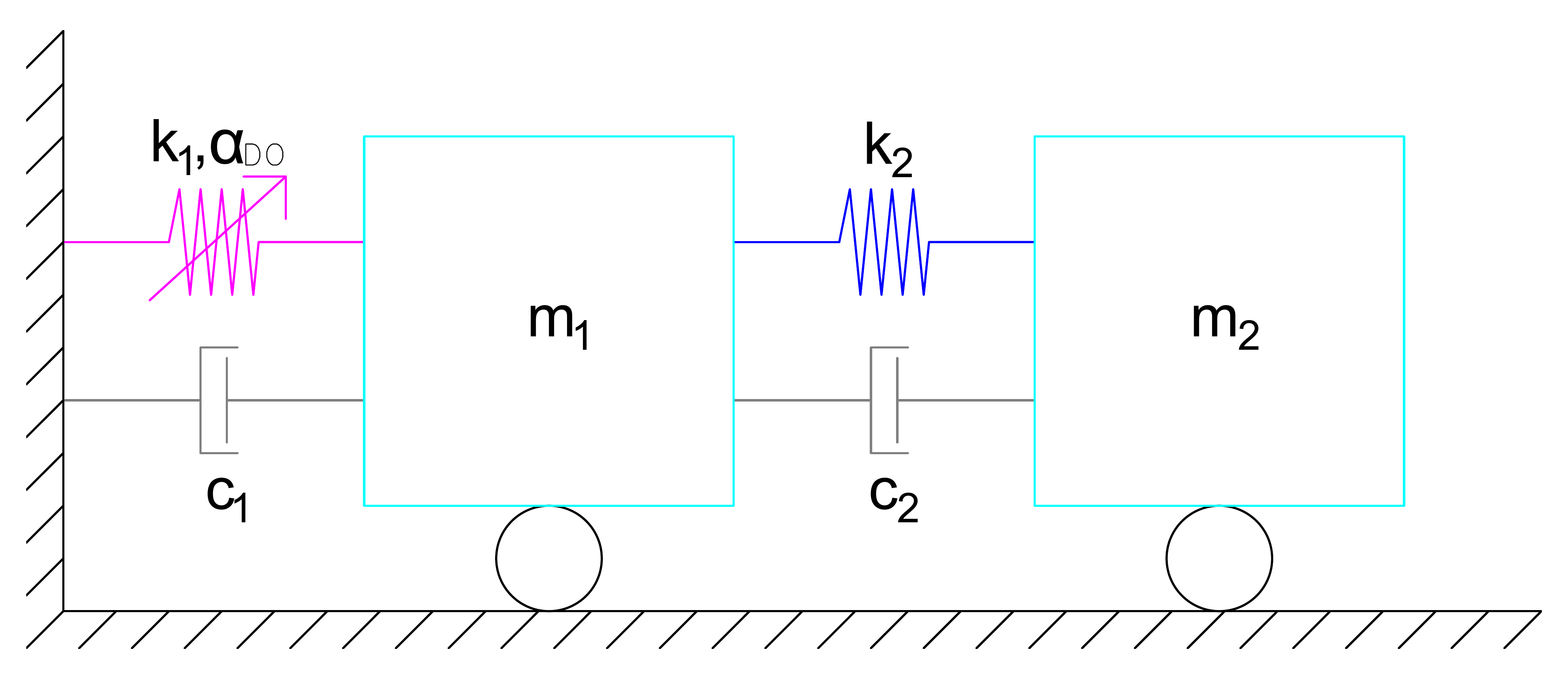}
    \caption{Schematic representation of the 2-DOF System with duffing oscillator considered in example 1. The nonlinear duffing oscillator is attached with the first degree of freedom (shown in magenta).}
    \label{fig:2dof}
\end{figure}

As the first example, we consider a 2-DOF system as shown in Fig. \ref{fig:2dof}.
The nonlinear duffing oscillator is attached with the first degree of freedom.
The coupled governing equations for this system are represented as
\begin{equation}\label{2dof-dynamic}
\begin{aligned}
&m_1\ddot{x}_1+c_1\dot{x}_1+k_1x_1+\alpha_{DO} {x_1}^3+c_2(\dot{x}_1-\dot{x}_2)+k_2(x_1-x_2)=\sigma_1\dot{W}_1+f_1, \\
&m_2\ddot{x}_2+c_2(\dot{x}_2-\dot{x}_1)+k_2(x_2-x_1)=\sigma_2\dot{W}_2+f_2,
\end{aligned}
\end{equation}
where $m_i$, $c_i$ and $k_i$, respectively, represent the mass, damping and stiffness of the $i-$th degree of freedom. Although not explicitly shown, it is to be noted that $k_i$ changes with the slow time-scale $t_s$. $F_i$ and $\sigma_i\dot{W}_i$, respectively, represents the deterministic and the stochastic force acting on the $i-$th floor. $\alpha_{DO}$ controls the nonlinearity in the system.
The parametric values considered for this example are shown in Table \ref{tab:param_eg1}.
\begin{table}[ht!]
 \caption{System Parameters for 2-DOF System}\label{tab:param_eg1}
    \centering
    \resizebox{1\linewidth}{!}{
    \begin{tabular}{|c|c|c|c|c|}
    \hline
    Mass&Stiffness&Damping&Force(N)&Stochastic Noise\\
    (Kg)&Constant (N/m)&Constant (Ns/m)&$F_i=\lambda_i sin(\omega_i t)$&Parameters\\
    \hline
    $m_1 = 20$&$k_1 = 1000$&$c_1 = 10$&$\lambda_1 = 10,\,\,\omega_1 = 10$&$s_1 = 0.1$\\
    $m_2 = 10$&$k_2 = 500$&$c_2 = 5$&$\lambda_2 = 10,\,\,\omega_2 = 10$&$s_2 = 0.1$\\
    \hline
    \multicolumn{5}{|c|}{DO Oscillator Constant, $\alpha_{DO}=100$}\\
    \hline
    \end{tabular}
    }
\end{table}

\begin{figure}[ht!]
    \centering
    \includegraphics[scale = 0.35]{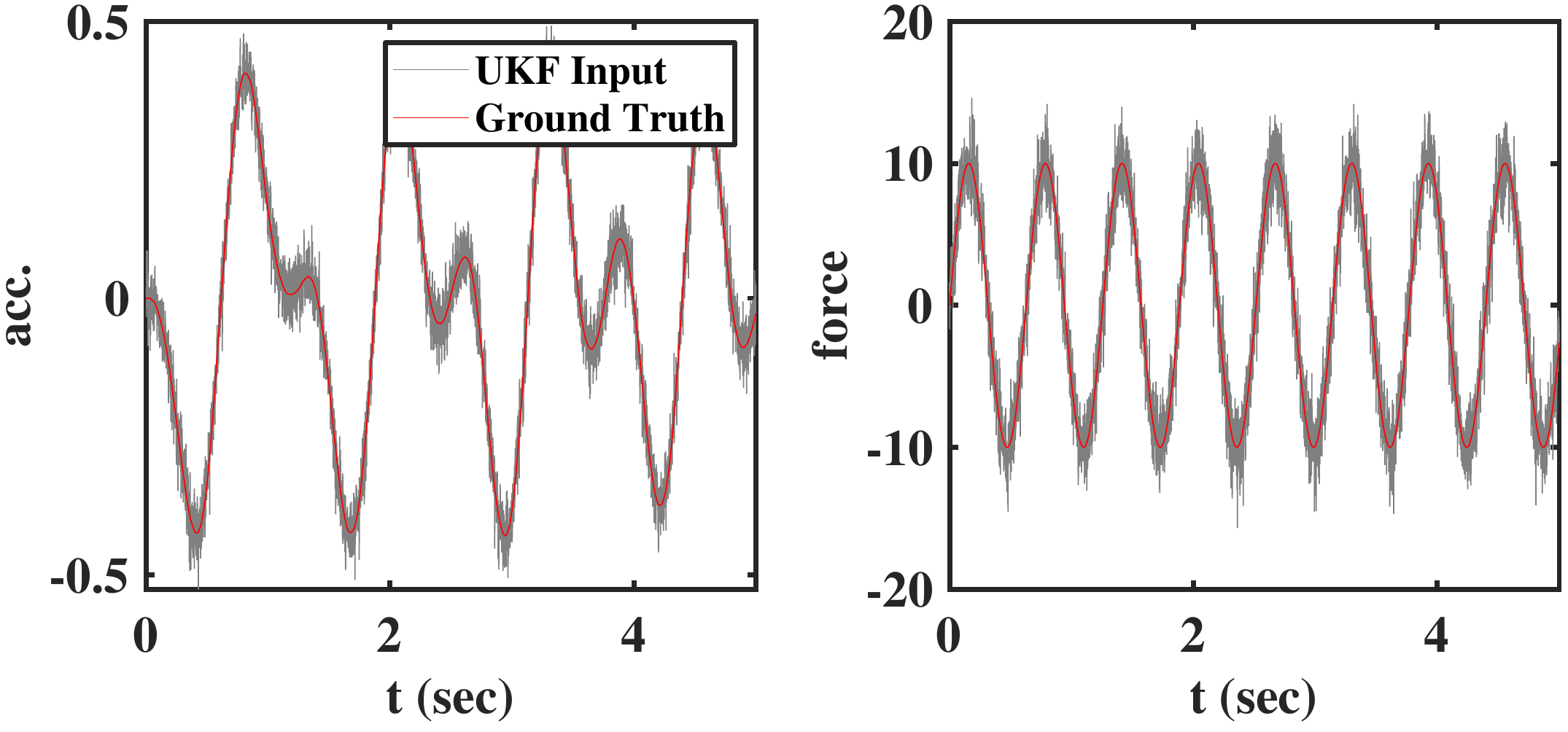}
    \caption{Sample Acceleration and deterministic component of the force for the 2-DOF problem. The stochasticity observed for the force is due to the noise present. Note that there is an additional stochastic component of force as shown in Eq. (\ref{eq:em}).}\label{fig:samp_f_acc_2dof}
\end{figure}

\begin{figure}[ht!]
    \centering
    \includegraphics[scale = 0.35]{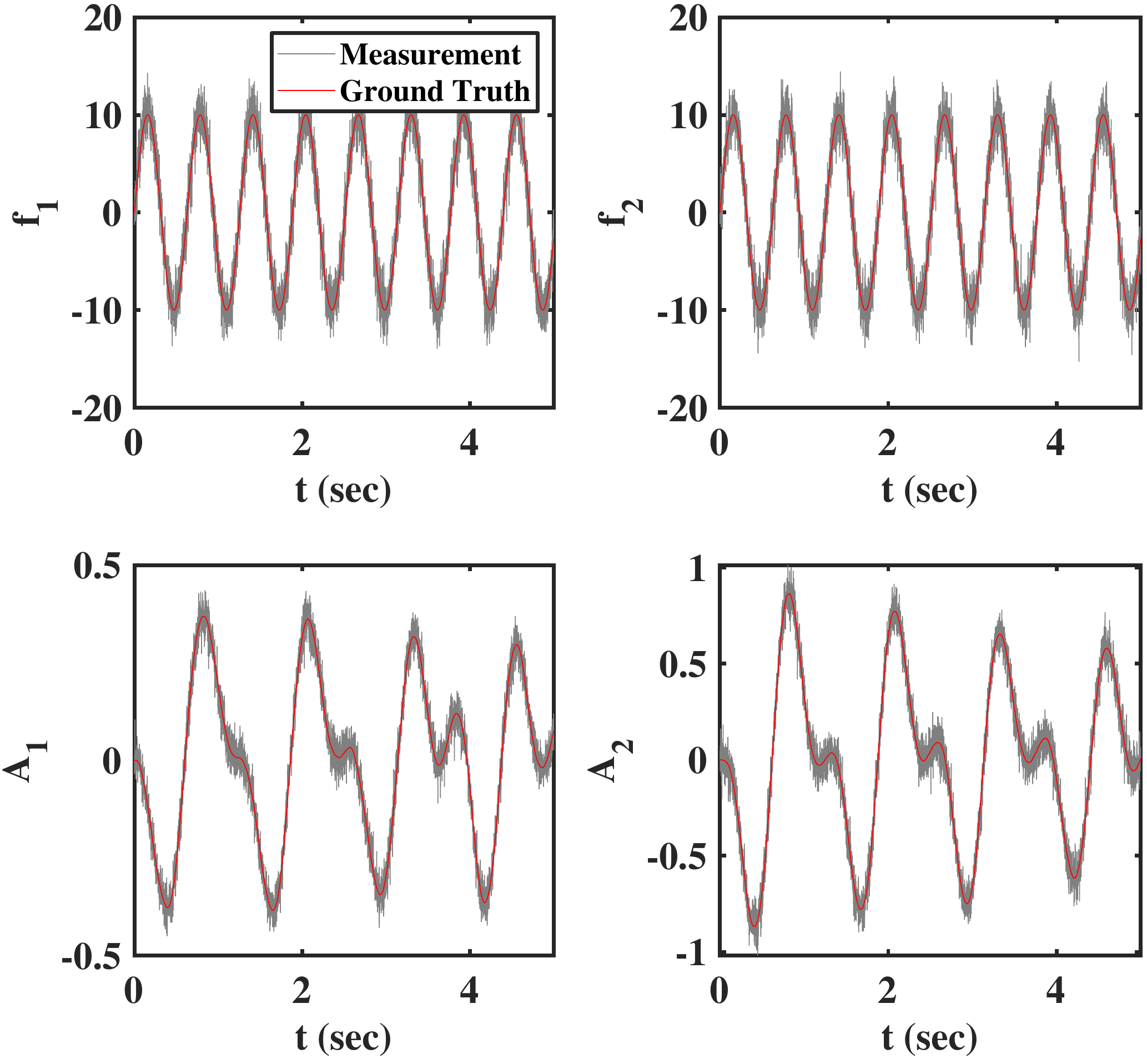}
    \caption{ Deterministic component of force and acceleration vectors at the two DOFs. The noisy acceleration vectors are provided as measurement to the UKF model.}
    \label{fig:fa-2dof-1dp}
\end{figure}

\begin{figure}[ht!]
\begin{subfigure}{1\textwidth}
    \centering
    \includegraphics[scale = 0.35]{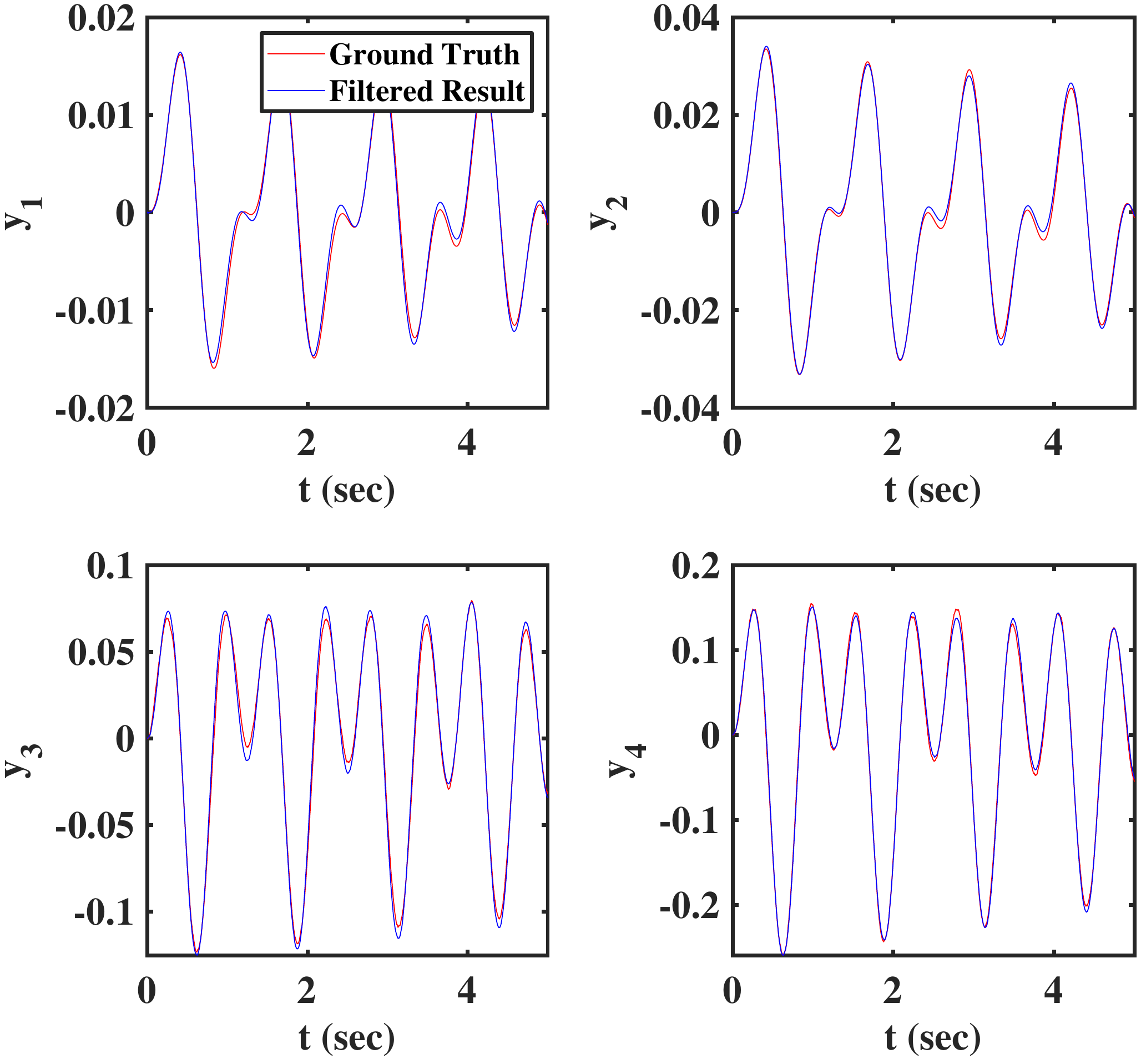}
    \caption{ State (Displacement And Velocity) Estimation}
\end{subfigure}
\begin{subfigure}{1\textwidth}
    \centering
    \includegraphics[scale = 0.35]{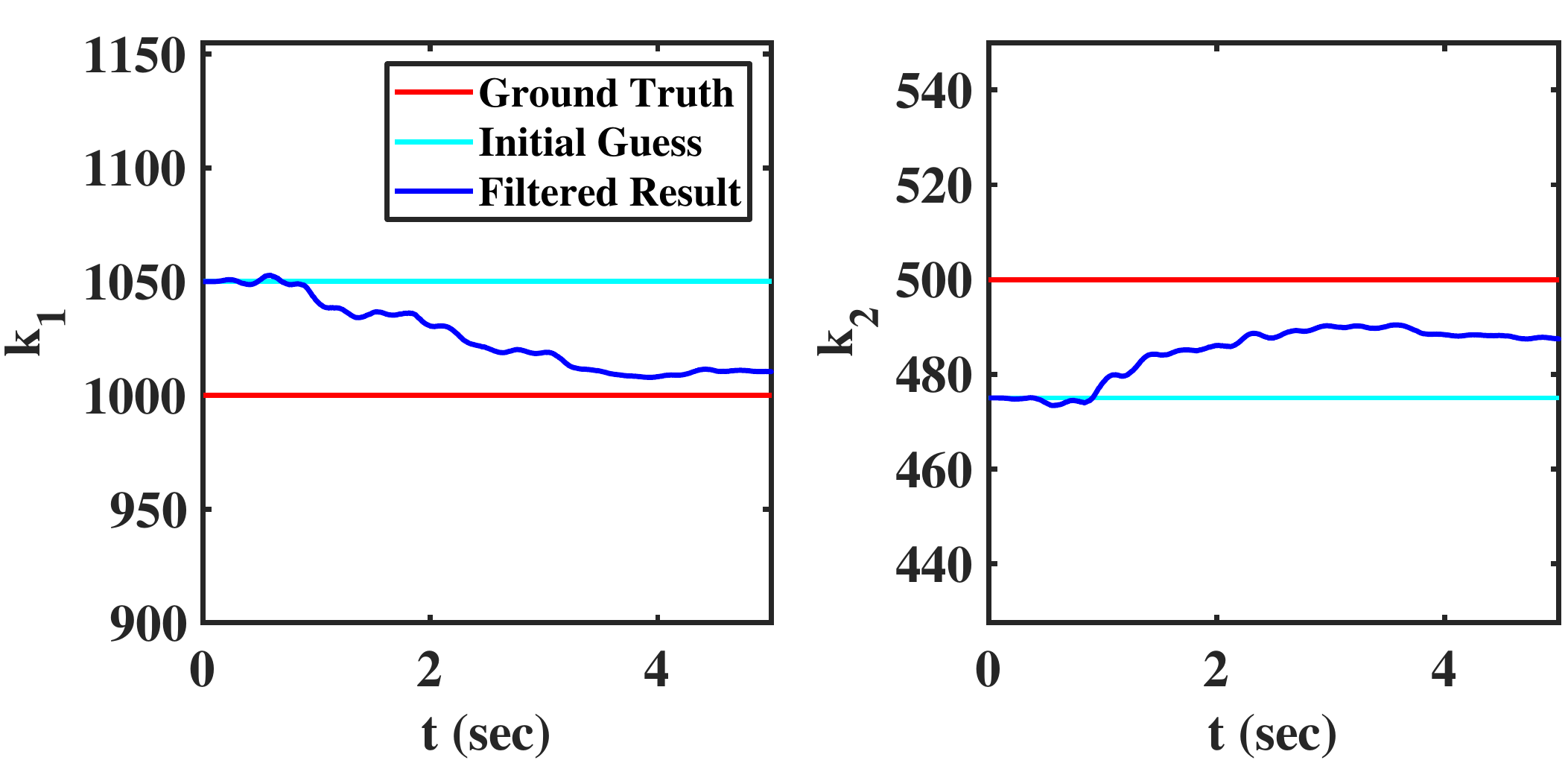}
    \caption{ Parameter (Stiffness) Estimation}
\end{subfigure}
\caption{ Combined state and parameter estimation results for the 2DOF system. Noisy measurements of acceleration at both the DOFs are provided as input to the UKF algorithm. The results corresponds to the initial measurement data.}
\label{fig:2dof-1dp}
\end{figure}

\begin{figure}[ht!]
\begin{subfigure}{1\textwidth}
    \centering
    \includegraphics[scale = 0.35]{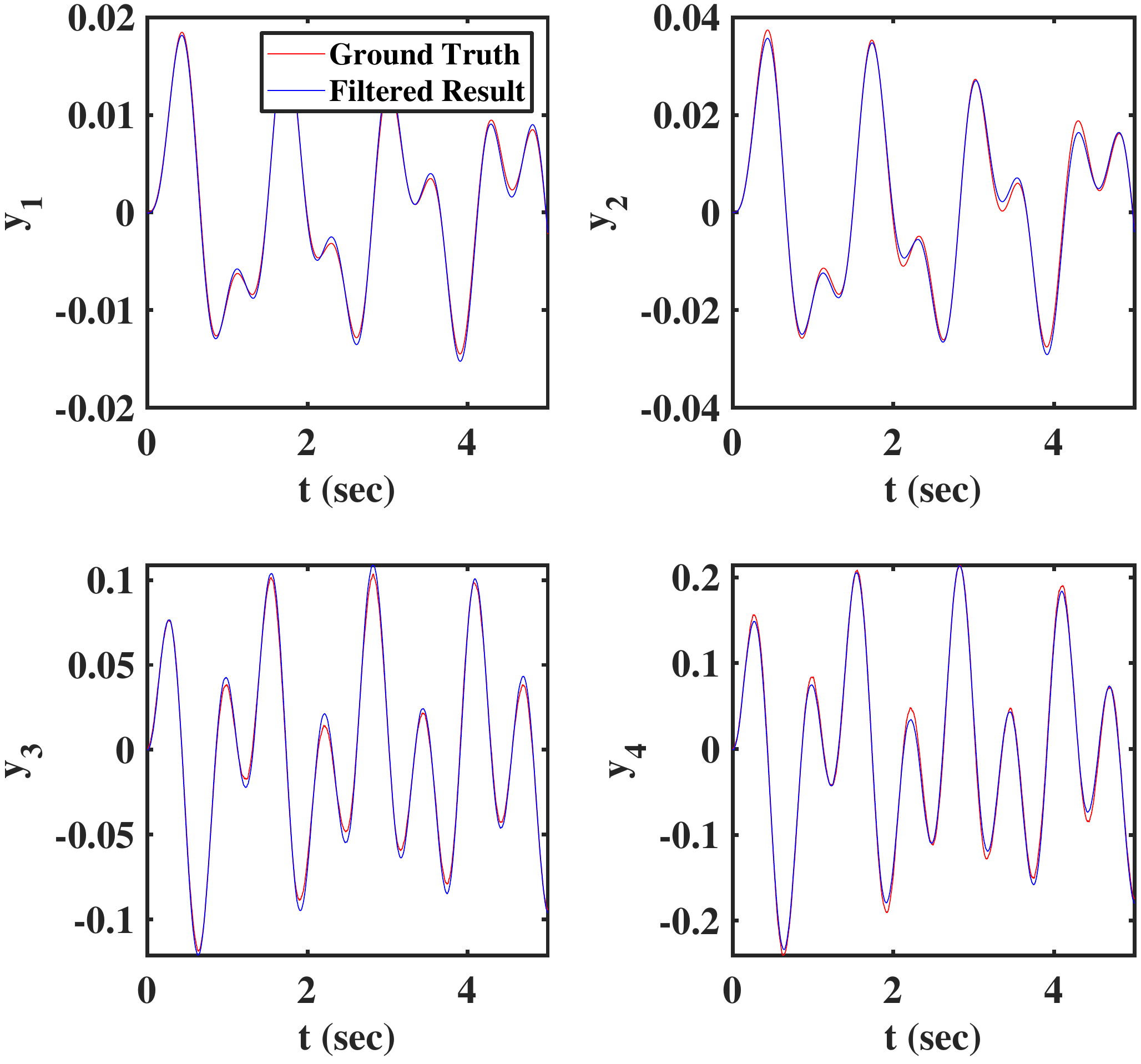}
    \caption{ State (Displacement And Velocity) Estimation}
\end{subfigure}
\begin{subfigure}{1\textwidth}
    \centering
    \includegraphics[scale = 0.35]{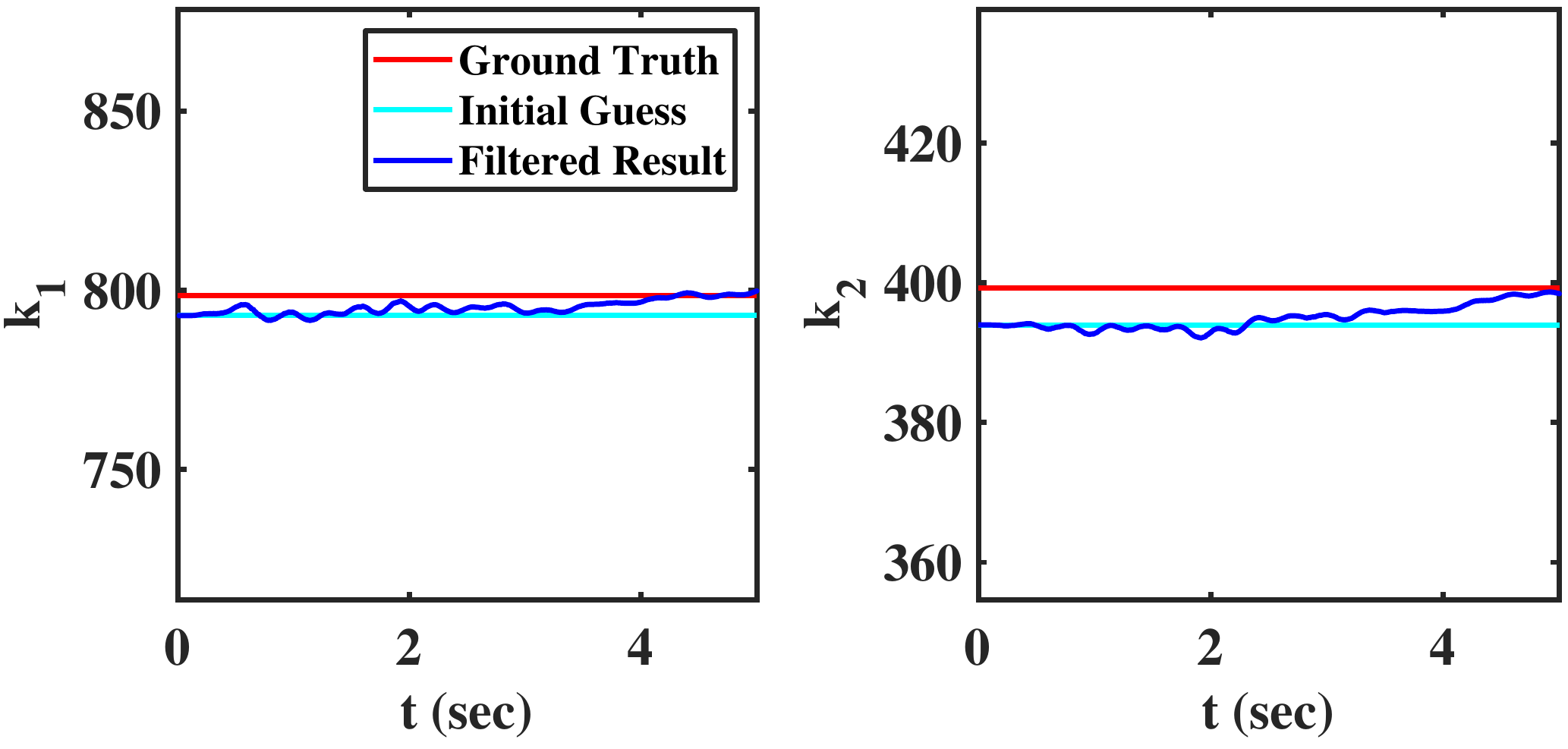}
    \caption{ Parameter (Stiffness) Estimation}
\end{subfigure}
\caption{ Combined state and parameter estimation results for the 2DOF system. Noisy measurements of acceleration at both the DOFs are provided as input to the UKF algorithm. The results corresponds to an intermediate measurement data.}
\label{2dof-1dp-91}
\end{figure}

\begin{figure}[ht!]
    \centering
    \includegraphics[scale = 0.35]{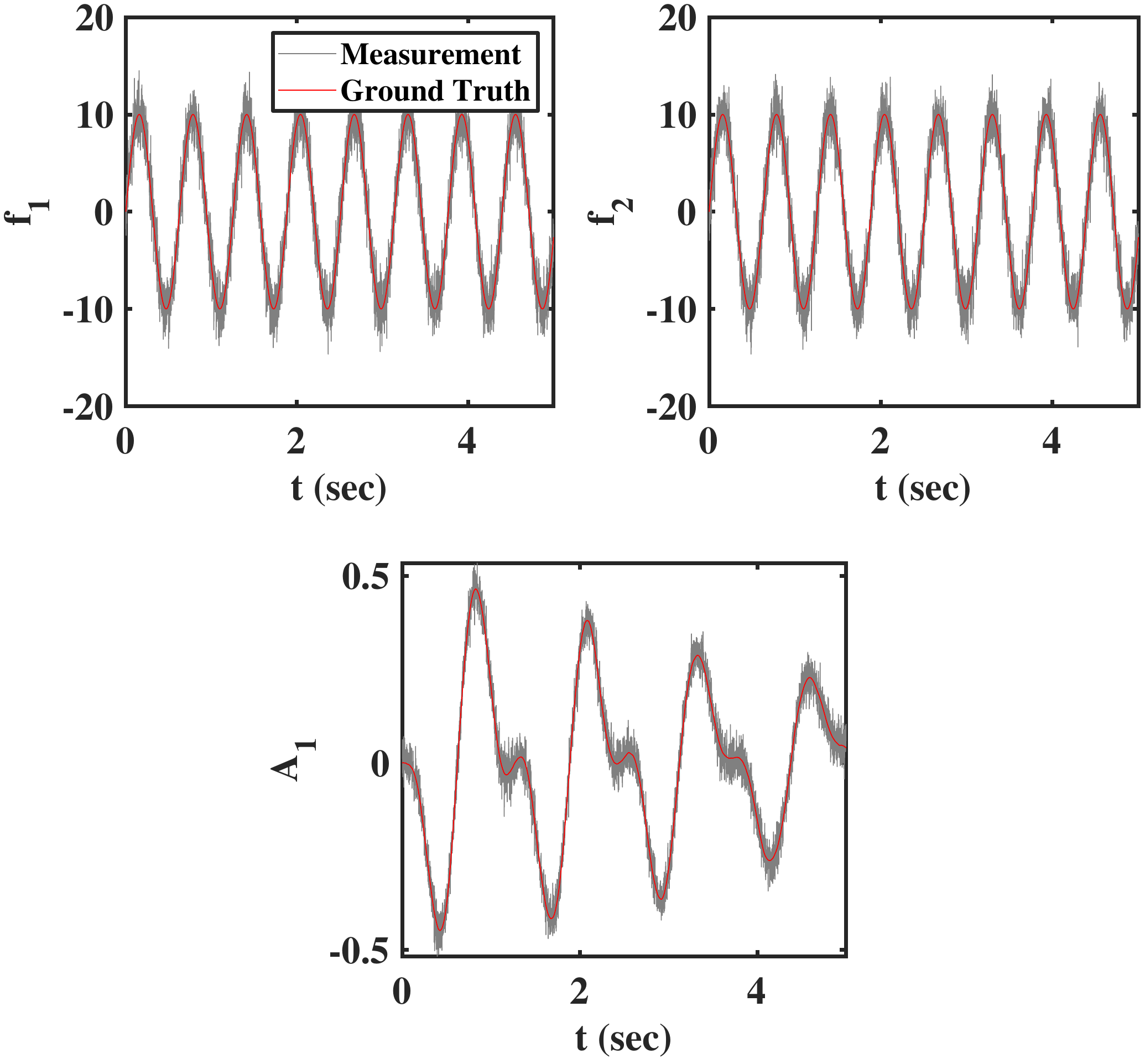}
    \caption{Deterministic component of force and acceleration vector used in UKF. For this case, only acceleration measurements at first degree of freedom is available.}
    \label{fa-2dof-1dp-1acc}
\end{figure}

\begin{figure}[ht!]
\begin{subfigure}{1\textwidth}
    \centering
    \includegraphics[scale = 0.35]{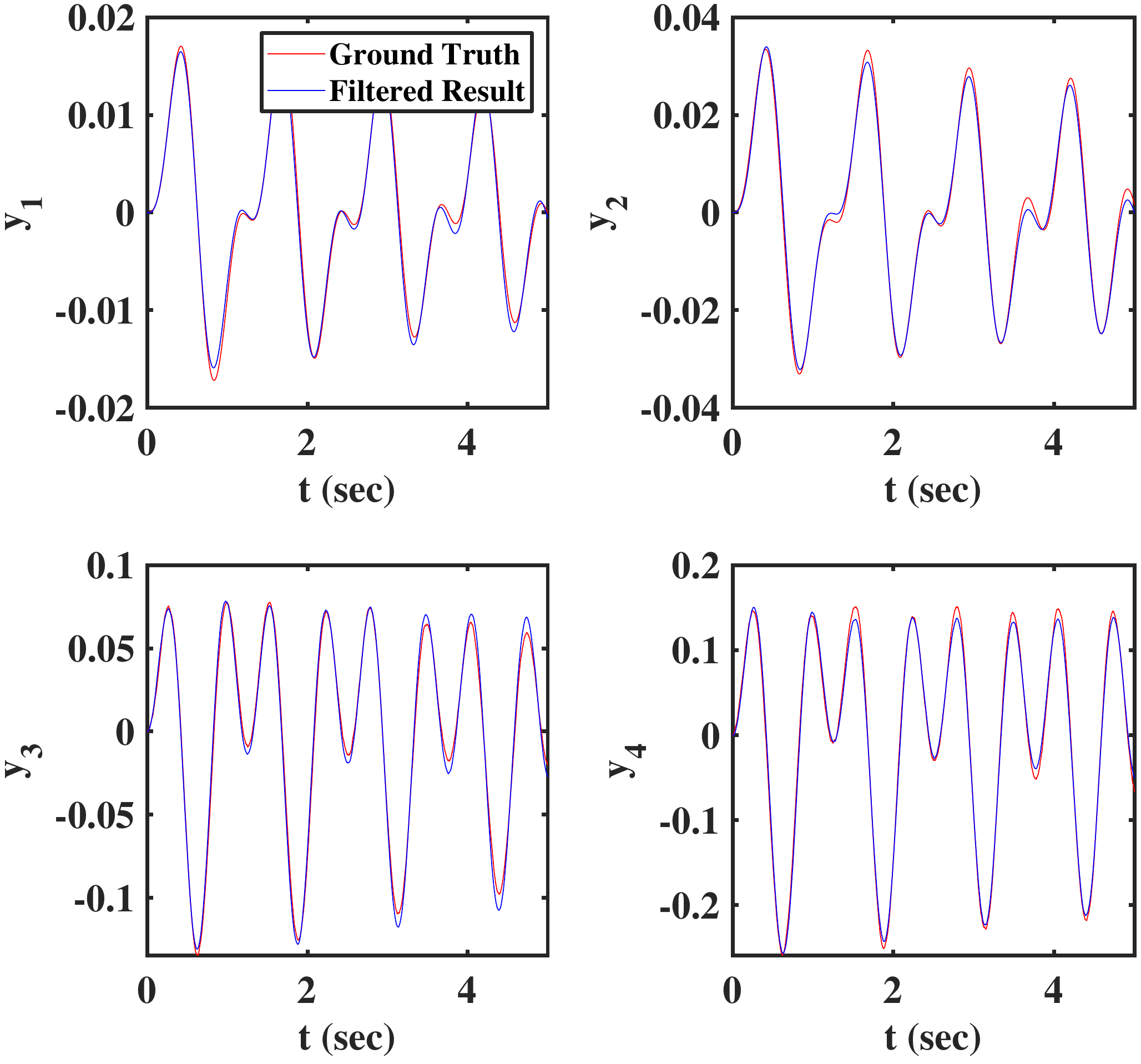}
    \caption{State (Displacement And Velocity) Estimation}
\end{subfigure}
\begin{subfigure}{1\textwidth}
    \centering
    \includegraphics[scale = 0.35]{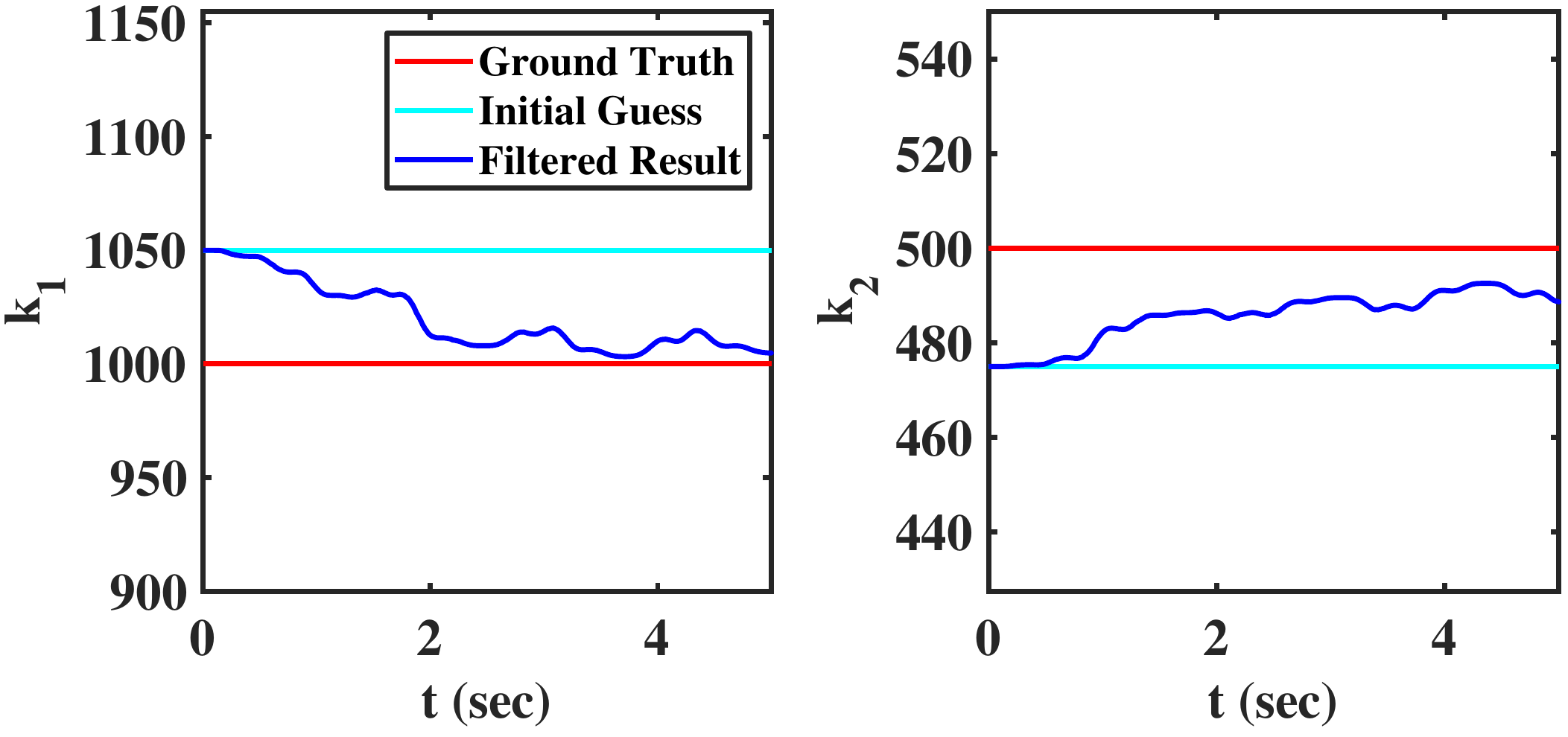}
    \caption{Parameter (Stiffness) Estimation}
\end{subfigure}
\caption{Combined state and parameter estimation results for the 2DOF system estimated from only one acceleration measurements. The noisy acceleration measurement at DOF 1 is provided to the UKF as measurement.}
\label{2dof-1dp-1acc}
\end{figure}

\begin{figure}[ht!]
\centering
\begin{subfigure}{0.45\textwidth}
    \centering
    \includegraphics[width=\textwidth]{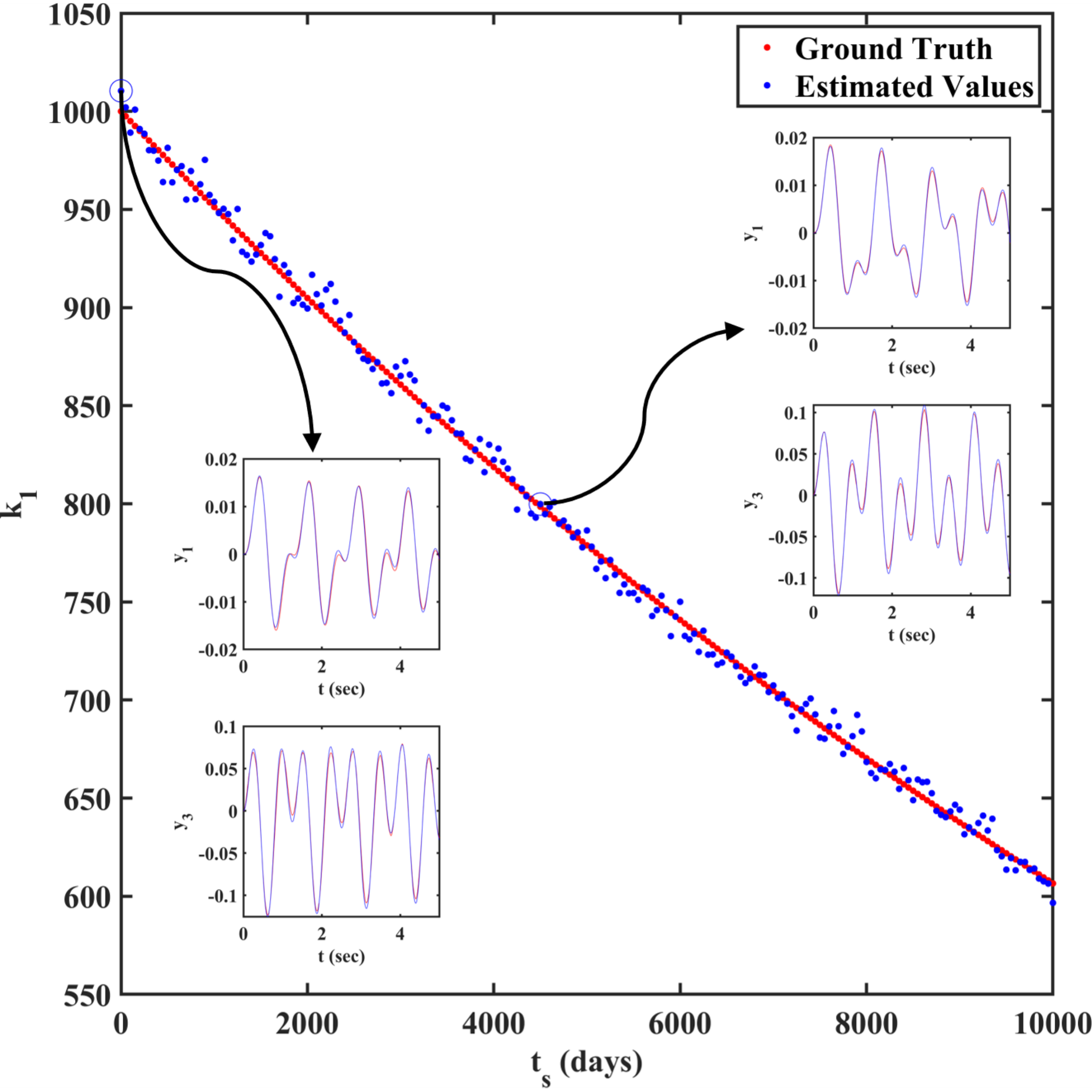}
    \caption{Stiffness ($k_1$)}
\end{subfigure}
\begin{subfigure}{0.45\textwidth}
    \centering
    \includegraphics[width=\textwidth]{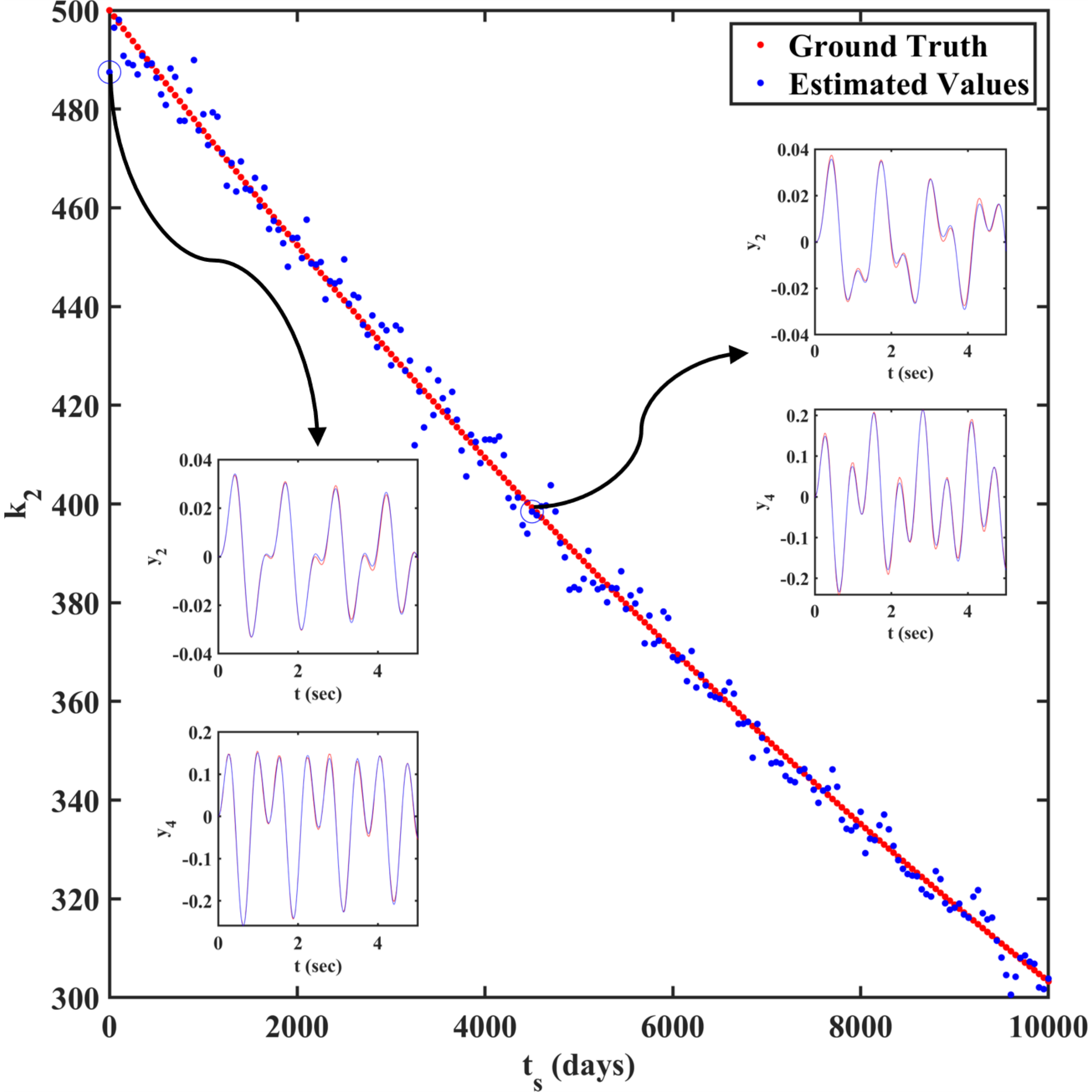}
    \caption{Stiffness ($k_2$)}
\end{subfigure}
\caption{Estimated stiffness in slow-time-scale using the UKF algorithm for the 2DOF example. State estimations at selected time-steps are also shown. Good match between the ground truth and the filtered result is obtained. These data act as input to the Gaussian process (GP).}
\label{f2}
\end{figure}

\begin{figure}[ht!]
    \centering
    \includegraphics[width=0.8\textwidth]{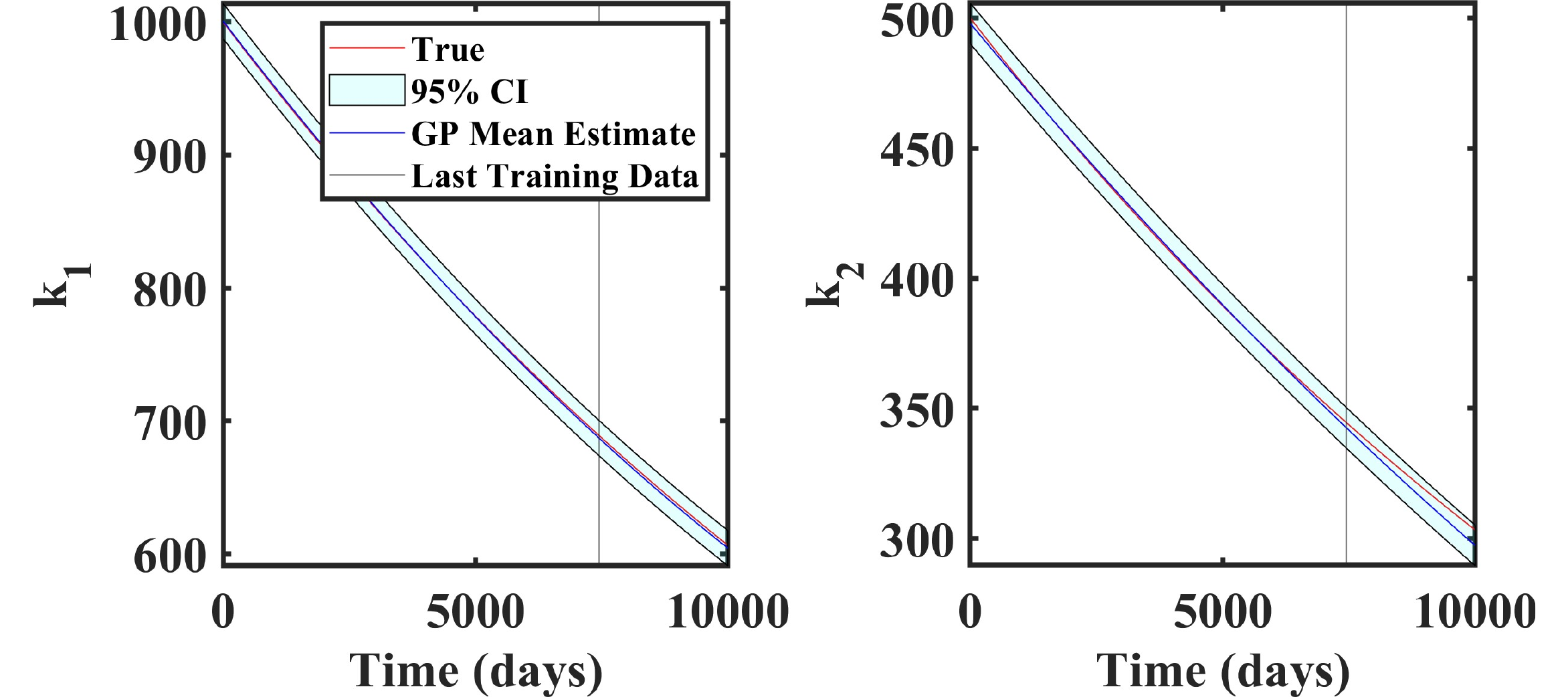}
    \caption{Results representing the performance of the proposed digital twin for the 2DOF system. The GP is trained using the data generated using UKF. Data upto the horizontal line is available to the GP. The digital twin performs well even when predicting system parameters at future time-steps.}
    \label{gp2}
\end{figure}

\noindent The system states are defined as:
\begin{equation}
    \begin{matrix}x_1 = y_1, & x_2 = y_2,\\
    \dot{x}_1 = y_3, & \dot{x}_2 = y_4,\end{matrix}
\end{equation}
and the governing equation in Eq. (\ref{2dof-dynamic}) is represented in the form of Ito-diffusion equations to obtain the drift and dispersion coefficients:
\begin{equation}\label{eq:ito}
d \bm y= \bm a\,dt + \mathbf b\,d\bm W,
\end{equation}
where
\begin{subequations}
\begin{equation}
    \bm a=\left[\begin{array}{c} y_{3}\\ y_{4}\\ \frac{f_1}{m_1}-\frac{1}{m_1}\left(c_{1}\,y_{3}+c_{2}\,y_{3}-c_{2}\,y_{4}+k_{1}\,y_{1}+k_{2}\,y_{1}-k_{2}\,y_{2}+\alpha_{do} \,{y_{1}}^3\right)\\ \frac{1}{m_2}\left({c_{2}\,y_{3}-c_{2}\,y_{4}+k_{2}\,y_{1}-k_{2}\,y_{2}}{m_{2}}\right)+\frac{f_2}{m_2} \end{array}\right]
\end{equation}
\begin{equation}
    \mathbf{b}=\left[\begin{matrix}0 & 0 \\ 0 & 0\\ \frac{{\sigma}_{1}}{m_{1}} & 0\\ 0 & \frac{{\sigma}_{2}}{m_{2}}\end{matrix}\right].
\end{equation}
\end{subequations}
For illustrating the performance of the proposed digital twin, we generate synthetic data by simulating Eq. (\ref{eq:ito}).
The data simulation is carried out using Taylor 1.5 strong scheme \cite{tripura2020ito,roy2017stochastic}
\begin{equation}
\begin{aligned}
\bm y_{k+1}=(\bm y + \bm a \Delta t+&\mathbf{b}\Delta \bm w+0.5L^j(\mathbf{b}) \left( {\Delta {w^2} - \Delta t} \right) + L^j(\bm a)\Delta \bm z\\
&+L^0(\mathbf{b}) \left( {\Delta {w}\Delta t - \Delta {z}} \right) +0.5L^0(\bm a){\Delta t}^2)_k
\label{eq:T1.5}
\end{aligned}
\end{equation}
where, $L^0$ and $L^j$ are Kolmogorov operators \cite{roy2017stochastic} evaluated on drift and diffusion coefficients i.e. on elements of $\bm a$ and $\textbf{b}$. $\bm \Delta w$ and $\bm \Delta z$ are the Brownian increments \cite{roy2017stochastic} evaluated at each time step $\Delta t$.
Before proceeding with the performance of the proposed DT, we investigate the performance of UKF in joint parameter state estimation.
To avoid the so called `inverse crime' \cite{wirgin2004inverse}, Euler Maruyama (EM) integration scheme is used during filtering.
\begin{equation}\label{eq:em}
    \bm y_{k+1} = \left( \bm y + \bm a \Delta t + \mathbf b \Delta \bm w \right)_k.
\end{equation}
It may be noted that EM integration scheme provides a lower-order approximation as compared to Taylor's 1.5 strong integration scheme. 
In other words, the data is generated using a more accurate scheme as compared to the filtering. This helps in emulating a realistic scenario. For combined state-parameter estimation, the state space vector is modified as $
\bm {y}=\left[ y_{1}\,\, y_{2} \,\, y_{3} \,\, y_{4} \,\, k_{1} \,\, k_{2} \right]^T$.
Consequently, $\bm a$ and $\mathbf b$ are also modified as:
\begin{subequations}
\begin{equation}
    \bm a=\left[\begin{array}{c} y_{3}\\ y_{4}\\ \frac{f_1}{m_1}-\frac{1}{m_1}\left({c_{1}\,y_{3}+c_{2}\,y_{3}-c_{2}\,y_{4}+k_{1}\,y_{1}+k_{2}\,y_{1}-k_{2}\,y_{2}+\alpha_{do} \,{y_{1}}^3}\right)\\ \frac{1}{m_2}\left({c_{2}\,y_{3}-c_{2}\,y_{4}+k_{2}\,y_{1}-k_{2}\,y_{2}}\right)+\frac{f_2}{m_2}\\0\\0\end{array}\right],
\end{equation}
\begin{equation}
    \mathbf b=\left[\begin{matrix} 0 &0\\ 0 &0\\ \frac{\sigma_1}{m_1} &0\\  0 &\frac{\sigma_2}{m_2}\\0 &0\\0 &0\end{matrix}\right].
\end{equation}
\end{subequations}
For obtaining the dynamic model function for UKF model, first two terms of EM algorithm are used.
\begin{equation}
\bm f(\bm y)= \bm y + \bm a \Delta t.
\label{EM1}
\end{equation}
For estimating the noise covariance $\mathbf Q$, $q$ is expressed as:
\begin{equation}
    \bm q = \mathbf q_c \bm {RV},
    \label{pnc:Q}
\end{equation}
where $\mathbf q_c$ is a constant diagonal matrix which is multiplied by vector of random variables $\bm {RV}$ to compute $\bm q$.
The basic form for $\mathbf q_c$ is extracted from the remaining terms of EM algorithm i.e., $\mathbf{b}\Delta \bm w$.
\begin{equation}
\begin{array}{c}
    \mathbf {q_c} = diag\left[\begin{array}{cccccc} 0& 0& \frac{\sigma_1\sqrt{dt}}{m_1}& \frac{\sigma_2\sqrt{dt}}{m_2}& 0& 0\end{array}\right],\\
    \mathbf Q = \mathbf q_c \mathbf q_c^T.
\end{array}
\end{equation}
The individual terms of $\mathbf Q$ can then be modified by any suitable factor to improve the accuracy of the filter.
Considering that the acceleration measurements are available to the DT, the simulated acceleration measurements are obtained as: follows:
\begin{equation}\label{eq:acc}
\bm A = -\mathbf{M}^{-1}(\bm G+\mathbf{K}X+\mathbf{C}\dot{X}),
\end{equation}
where $\mathbf M$, $\mathbf C$ and $\mathbf K$ are the mass, damping and stiffness matrices. $\bm G$, as already discussed in Eq. (\ref{eq:nominal}) is the contribution due to the nonlinearity in the system.
Eq. (\ref{eq:acc}) can be written in the state-space form as
\begin{equation}\label{eq:acc_2dof}
\bm A=\left[\begin{matrix}-\frac{1}{m_1}\left({c_{1}\,y_{3}+c_{2}\,y_{3}-c_{2}\,y_{4}+k_{1}\,y_{1}+k_{2}\,y_{1}-k_{2}\,y_{2}+\alpha_{do} \,{y_{1}}^3}\right)\\ \frac{1}{m_2}\left({c_{2}\,y_{3}-c_{2}\,y_{4}+k_{2}\,y_{1}-k_{2}\,y_{2}}\right)\end{matrix}\right].
\end{equation}
Using Eq. \ref{eq:acc_2dof}, the observation/measurement model for the UKF can be written as
\begin{equation}
\bm h(\bm y)=\left[\begin{matrix}-\frac{1}{m_1}\left({c_{1}\,y_{3}+c_{2}\,y_{3}-c_{2}\,y_{4}+k_{1}\,y_{1}+k_{2}\,y_{1}-k_{2}\,y_{2}+\alpha_{do} \,{y_{1}}^3}\right)\\ \frac{1}{m_2}\left({c_{2}\,y_{3}-c_{2}\,y_{4}+k_{2}\,y_{1}-k_{2}\,y_{2}}\right)\end{matrix}\right].
\end{equation}
The simulated acceleration measurements are corrupted by white Gaussian noise having a signal-to-noise ratio (SNR) of 50, where SNR is defined as: $\text{SNR} = {{\sigma _{\text{signal}}^2} \mathord{\left/
 {\vphantom {{\sigma _{data}^2} {\sigma _{noise}^2}}} \right.
 \kern-\nulldelimiterspace} {\sigma _{\text{noise}}^2}}$ and $\sigma$ is the standard deviation.
The deterministic force vector is also corrupted by white Gaussian noise having SNR of 20. 
Representative examples of acceleration and deterministic force for this problem are shown in Fig. \ref{fig:samp_f_acc_2dof}.
We use UKF along with the acceleration and deterministic force measurements, $\bm f(\bm y)$ and $\bm h(\bm y)$ for combined parameter state estimation.
Fig. \ref{fig:2dof-1dp} shows the combined state parameter estimation results for first data point i.e. $t_{s(i)} = t_{s(1)}$ of 2- DOF system. 
This is a relatively simple case where measurements at both degrees of freedom are available (see Fig. \ref{fig:fa-2dof-1dp}).
It can be observed that UKF provides highly accurate estimates of the state vectors.
As for parameter estimation (see Fig. \ref{fig:2dof-1dp}(b)), we observe that UKF provides highly accurate estimate for $k_1$. As for $k_2$, compared to the ground truth ($k_2 = 500$N/m), the proposed approach ($k_2 = 487.5$N/m) provides an accuracy of around 98\%.
Fig. \ref{2dof-1dp-91} shows the result for an intermediate data point i.e. at time $t_{s(i)} = t_{s(91)}$. The initial values of parameter while filtering are taken as the final values of parameter obtained from previous data point. Similar to that observed for the initial data point, Fig. \ref{2dof-1dp-91} shows that the filter manages to estimate the states accurately and also improves upon the parameter estimation.
\begin{figure}[ht!]
    \centering
    \includegraphics[scale = 0.35]{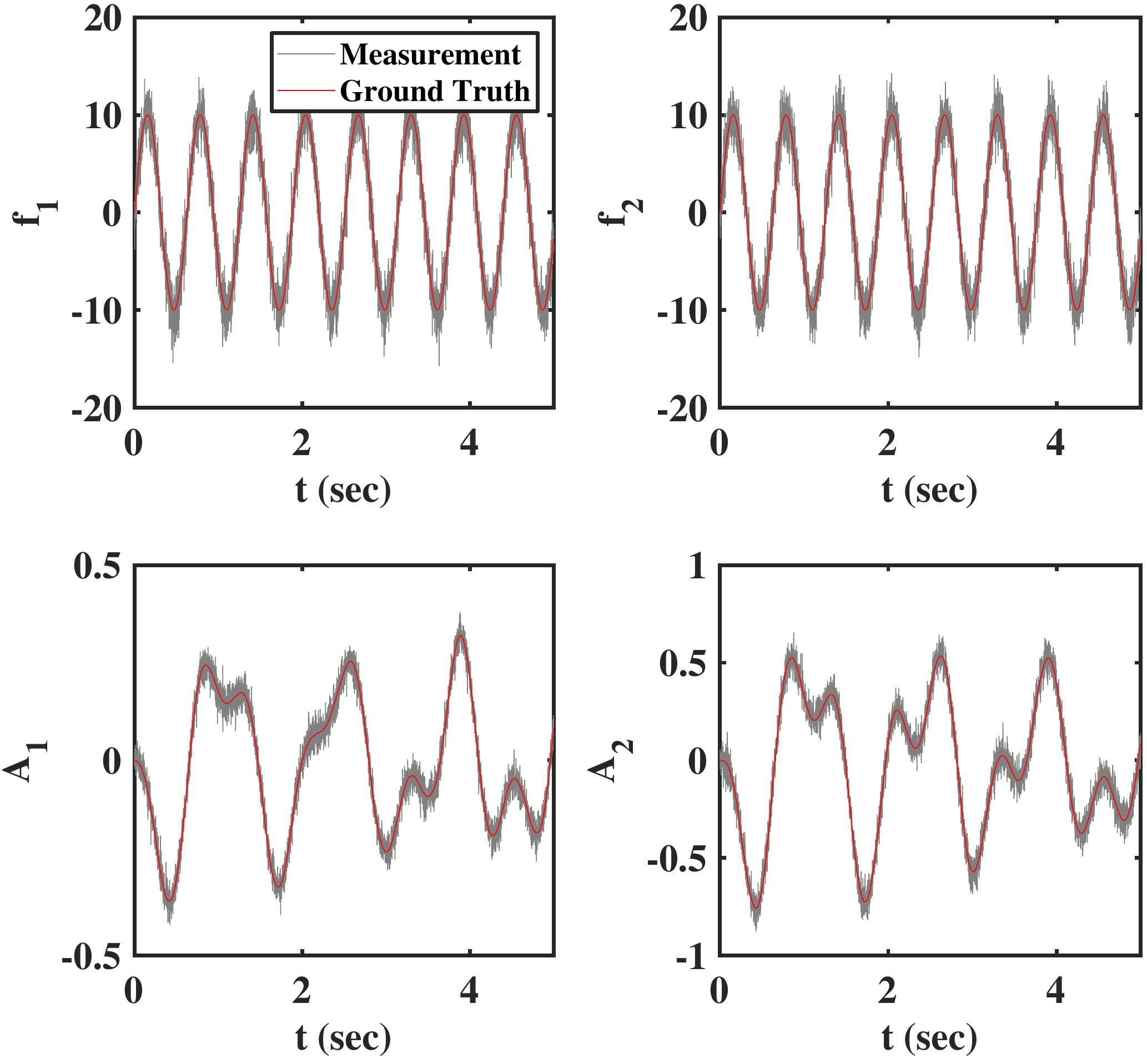}
    \caption{ Force (deterministic part) and acceleration vectors at an intermediate time-step. The noisy accelerations at the 2DOFs are provided as measurements to the UKF algorithm.}
\end{figure}
\noindent Next, we consider a more challenging scenario where data at only one DOF is available.
To be specific, acceleration measurements at DOF-1 is considered to be available (see Fig. \ref{fa-2dof-1dp-1acc}).
This changes the measurement model $h(.)$ while filtering and reduce it to,
\begin{equation}
h(y) = -\frac{1}{m_1}\left({c_{1}\,y_{3}+c_{2}\,y_{3}-c_{2}\,y_{4}+k_{1}\,y_{1}+k_{2}\,y_{1}-k_{2}\,y_{2}+\alpha_{do} \,{y_{1}}^3}\right)  .  
\end{equation}
Fig. \ref{2dof-1dp-1acc} shows the state and parameter estimation results for this case.
Similar to previous case, it can be observed that the state estimate and the estimate for $k_1$ are obtained with high degree of accuracy (see Fig. \ref{2dof-1dp-1acc}). Estimate for $k_2$ also approaches the ground truth ($k_2 = 500N/m$) giving an accuracy of approximately 98\%.\noindent
Finally, we focus on the other objective of the DT, which is to compute the time-evolution of the parameters, considering the stiffness to vary with slow time-scale $t_s$ as follows.
\begin{equation}\label{eq:degradation}
    k(t_s) = k_0\delta,
\end{equation}
where
\begin{equation}
    \delta = e^{-0.5\times10^{-4}\times t_s}.
    \label{del}
\end{equation}
We consider that acceleration measurements are available for 5 seconds every 50 days.
UKF is utilized as discussed before for computing the stiffness at each time-steps.
The resulting data obtained is shown in Fig. \ref{f2}.
Once the data points are obtained, GP is employed to evaluate the temporal evolution of the parameters.
Fig. \ref{gp2} shows results obtained using GP. 
The vertical lines in Fig. \ref{gp2} indicate
the time until which data is provided to the GP.
It is observed that GP yields highly accurate estimate of the two stiffness. 
Interestingly, results obtained using GP are not only accurate in the time-window (indicated by the vertical line) but also outside.
This indicates the the proposed DT can be used for predicting the system parameters at future time-step which in-turn can be used for predicting the future responses and solving remaining useful life and
predictive maintenance optimization problems.
Additionally, GP being a Bayesian machine learning algorithm provides an estimate of the confidence interval.
This can be used for collecting more data and in decision making.
\subsection{7-DOF system with duffing van der pol oscillator}

\begin{figure}[ht!]
    \centering
    \includegraphics[scale = 0.35]{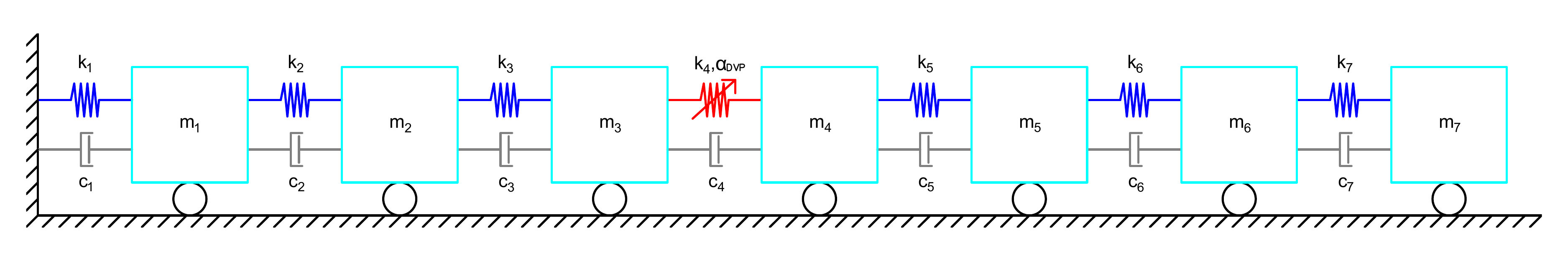}
    \caption{Schematic representation of the 7-DOF System with duffing Van-der Pol oscillator considered in example 2. The nonlinear DVP oscillator is attached with the fourth degree of freedom (shown in red).}
    \label{7dof}
\end{figure}

As our second example we consider a 7-DOF system as shown in Fig. \ref{7dof}. The 7-DOF system is modeled with a DVP oscillator at fourth DOF. The governing equations of motion for the 7-DOF system are given as,

\begin{equation}\label{7dof-dynamic}
    \textbf M \ddot {\bm X}  + \mathbf C \dot {\bm X}  + \mathbf K \bm X + \bm G \left( \bm X, \bm \alpha \right) = \bm F + \bm \Sigma {\bm {\dot{W}}}, 
\end{equation}
where 
$\mathbf M = diag\left[m_1,\ldots, m_7\right] \in \mathbb R^{7\times 7}$, $\bm X = \left[x_1,\ldots, x_7\right]^T \in \mathbb R^{7}$, $\bm \Sigma = diag\left[\sigma_1, \ldots, \sigma_7\right] \in \mathbb R^{7\times 7}$, 
$\bm {\dot{W}} = \left[\dot{W}_1, \ldots, \dot{W}_7\right]^T \in \mathbb R^{7}$, $\bm F = \left[f_1, \ldots, f_7\right]^T \in \mathbb R^{7}$ and
\[\bm G = \alpha_{DVP}\left[\begin{matrix}\bm 0_{1\times3}& (x_3-x_4)^3& (x_4-x_3)^3& \bm 0_{1\times2}\end{matrix}\right]^T \in \mathbb R^7.\]
$\mathbf C$ and $\mathbf K$ in Eq. (\ref{7dof-dynamic})
are tri-diagonal matrices representing damping and stiffness (linear component),
\begin{equation}
\mathbf C = \left[\begin{matrix}c_1+c_2 & -c_2 &  &  &  &  & \\
-c_2 & c_2+c_3 & -c_3 &  &  &  & \\
 & -c_3 & c_3+c_4 & -c_4 &  &  & \\
 &  & -c_4 & c_4+c_5 & -c_5 &  & \\
 &  &  & -c_5 & c_5+c_6 & -c_6 & \\
 &  &  &  & -c_6 & c_6+c_7 & -c_7\\
 &  &  &  &  & -c_6 & c_7\end{matrix}\right],
\end{equation}
\begin{equation}
\mathbf K = \left[\begin{matrix}k_1+k_2 & -k_2 &  &  &  &  & \\
-k_2 & k_2+k_3 & -k_3 &  &  &  & \\
 & -k_3 & k_3-k_4 & k_4 &  &  & \\
 &  & k_4 & -k_4+k_5 & -k_5 &  & \\
 &  &  & -k_5 & k_5+k_6 & -k_6 & \\
 &  &  &  & -k_6 & k_6+k_7 & -k_7\\
 &  &  &  &  & -k_6 & k_7\end{matrix}\right],
\end{equation}
where
$m_i$, $c_i$ and $k_i$, respectively, represent the mass, damping and stiffness of the $i-$th degree of freedom.
We have considered the stiffness of all, but the fourth DOF to vary with the slow time-scale $t_s$.
The rationale behind not varying the stiffness corresponding to the 4th DOF resides in the fact that 
nonlinear stiffness is generally used for vibration control \cite{das2021robust} and energy harvesting \cite{cao2019novel} and hence, is kept constant.
Parametric values for the 7-DOF system are shown in Table \ref{7dof-t1}.
\begin{table}[ht!]
    \centering
    \resizebox{1\linewidth}{!}{
    \begin{tabular}{|c|c|c|c|c|c|}
    \hline
    Index&Mass&Stiffness&Damping&Force(N)&Stochastic Noise\\
    i&(Kg)&Constant (N/m)&Constant (Ns/m)&$F_i=\lambda_i sin(\omega_i t)$&Parameters\\
    \hline
    i = 1,2&$m_i = 20$&$k_i = 2000$&\multirow{3}{*}{$c_i = 20$}&\multirow{3}{*}{$\lambda_i = 10,\,\,\omega_i = 10$}&\multirow{3}{*}{$s_i = 0.1$}\\
    \cline{1-3}
    i = 3,4,5,6&$m_i = 10$&$k_i = 1000$&&&\\
    \cline{1-3}
    i = 7&$m_i = 5$&$k_i = 500$&&&\\
    \hline
    \multicolumn{6}{|c|}{DVP Oscillator Constant, $\alpha_{DVP}=100$}\\
    \hline
    \end{tabular}
    }
    \caption{System Parameters -- 7-DOF System -- Data Simulation}
    \label{7dof-t1}
\end{table}

\begin{figure}[ht!]
    \centering
    \includegraphics[scale = 0.4]{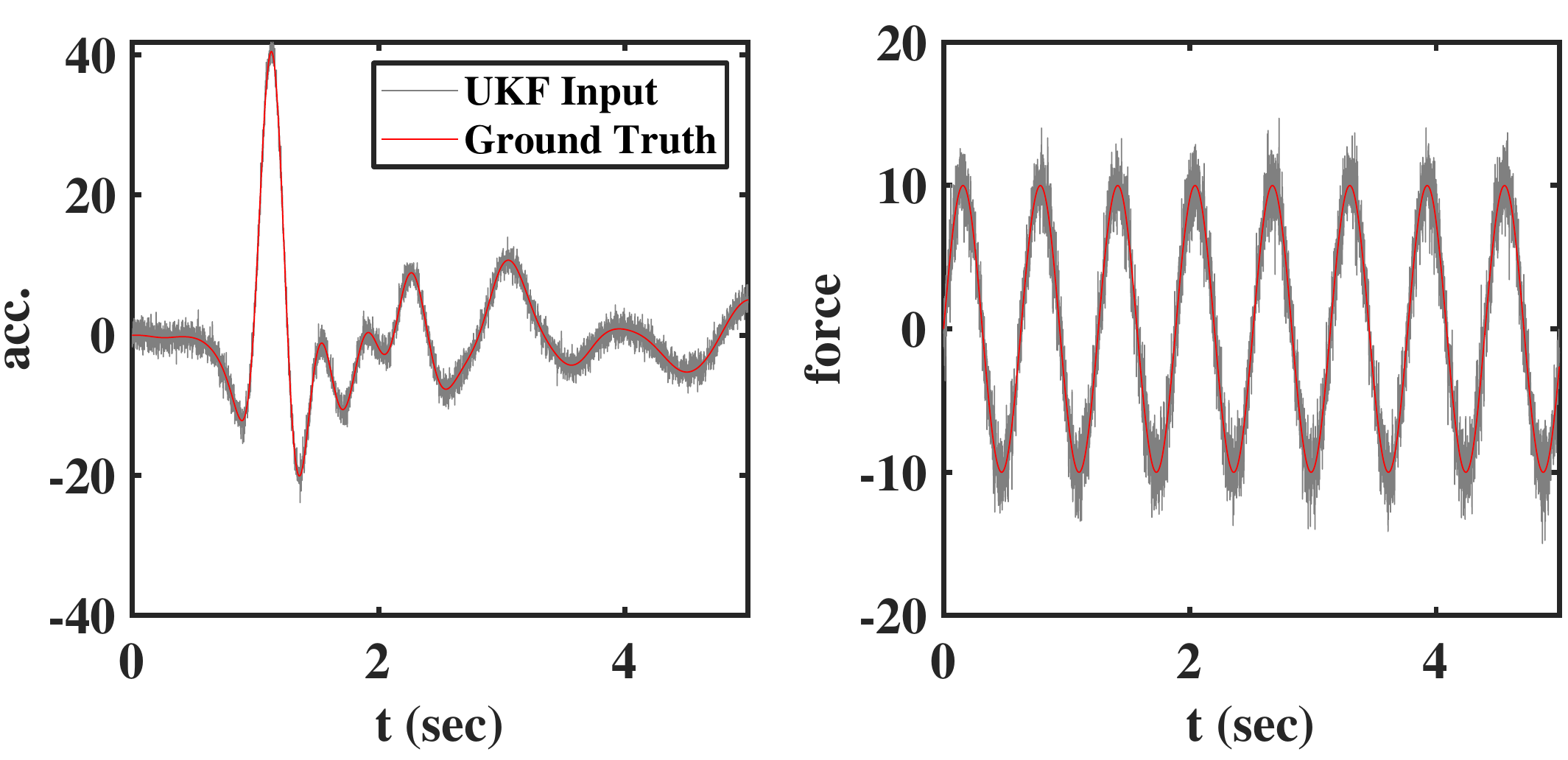}
    \caption{Sample Acceleration and deterministic component of the force for the 7-DOF problem. The stochasticity observed for the force is due to the noise present. Note that there is an additional stochastic component of force as shown in Eq. (\ref{eq:em}).}
    \label{acc-f-7dof}
\end{figure}

To convert the governing equations for 7 DOF system to state space equations, the following transformations are considered:
\begin{equation}
    \begin{matrix}x_1 = y_1, & \dot{x}_1 = y_2, & x_2 = y_3, & \dot{x}_2 = y_4, & x_3 = y_5, & \dot{x}_3 = y_6, & x_4 = y_7\\
    \dot{x}_4 = y_8, & x_5 = y_9, & \dot{x}_5 = y_{10}, & x_6 = y_{11}, & \dot{x}_6 = y_{12}, & x_7 = y_{13}, & \dot{x}_7 = y_{14}\end{matrix}
\end{equation}
Using Eq. (\ref{eq:ito}), dispersion and drift matrices for 7-DOF system are identified as follows:
\begin{equation}\label{7bm}
     b_{ij} = \left\{\begin{array}{ll}
     \frac{\sigma_i}{m_i},&\text{ for }i = 2j\text{ and }j=(1,2,3,5,6,7)\\
     \frac{\sigma_i}{m_i}y_{2j-1},&\text{ for }i = 2j\text{ and }j = 4\\
     0,&\text{ elsewhere}
     \end{array}\right.
 \end{equation}
\begin{equation}
\bm a=\left[\begin{array}{c} y_{2}\\
\frac{f_1}{m_1}-\frac{1}{m_1}\left({y_{1}\,\left(k_{1}+k_{2}\right)-c_{2}\,y_{4}-k_{2}\,y_{3}+y_{2}\,\left(c_{1}+c_{2}\right)}\right)\\
y_{4}\\
\frac{f_2}{m_2}+\frac{1}{m_2}\left({c_{2}\,y_{2}-y_{3}\,\left(k_{2}+k_{3}\right)+c_{3}\,y_{6}+k_{2}\,y_{1}+k_{3}\,y_{5}-y_{4}\,\left(c_{2}+c_{3}\right)}\right)\\
y_{6}\\
\frac{f_3}{m_3}-\frac{1}{m_3}\left({k_{4}\,y_{7}-c_{4}\,y_{8}-k_{3}\,y_{3}-c_{3}\,y_{4}+y_{5}\,\left(k_{3}-k_{4}\right)+\alpha_{DVP} \,{\left(y_{5}-y_{7}\right)}^3+y_{6}\,\left(c_{3}+c_{4}\right)}\right)\\
y_{8}\\
\frac{f_4}{m_4}+\frac{1}{m_4}\left({c_{4}\,y_{6}+c_{5}\,y_{10}-k_{4}\,y_{5}+k_{5}\,y_{9}+y_{7}\,\left(k_{4}-k_{5}\right)+\alpha_{DVP} \,{\left\{y_{5}-y_{7}\right\}}^3-y_{8}\,\left(c_{4}+c_{5}\right)}\right)\\
y_{10}\\
\frac{f_5}{m_5}+\frac{1}{m_5}\left({c_{5}\,y_{8}-y_{9}\,\left(k_{5}+k_{6}\right)+c_{6}\,y_{12}+k_{5}\,y_{7}+k_{6}\,y_{11}-y_{10}\,\left(c_{5}+c_{6}\right)}\right)\\
y_{12}\\
\frac{f_6}{m_6}+\frac{1}{m_6}\left({c_{6}\,y_{10}-y_{11}\,\left(k_{6}+k_{7}\right)+c_{7}\,y_{14}+k_{6}\,y_{9}+k_{7}\,y_{13}-y_{12}\,\left(c_{6}+c_{7}\right)}\right)\\
y_{14}\\
\frac{f_7}{m_7}+\frac{1}{m_7}\left({c_{7}\,y_{12}-c_{7}\,y_{14}+k_{7}\,y_{11}-k_{7}\,y_{13}}\right) \end{array}\right]
\label{7am}
\end{equation}
Similar to previous example data simulation is carried out using Taylor-1.5-Strong algorithm shown in Eq. (\ref{eq:T1.5}) and filtering model is formed using EM equation shown in Eq. (\ref{eq:em}). For performing combined state parameter estimation state vector is modified to:
\begin{equation}
    \bm y = \left[\bm y_{1:14}, \bm k_{1:7} \right]^T
\end{equation}
Consequently $\bm a$ and $\bm b$ matrices are changed as: $\bm a = [\bm a_{state}^T, \bm 0_{1\times 6}]^T$ and $\bm b = [\bm b_{state}^T,\,\,\bm 0_{7\times 7}]^T$ where $\bm a_{state}$ and $\bm b_{state}$ are equal to $\bm a$ and $\bm b$ from Eq. (\ref{7am}) and Eq. (\ref{7bm}) respectively.
Note that although the value $k_4$ is a-priori known, we have still considered it into the state vector.
It was observed that such a setup helps in regularizing the UKF estimates.
Dynamic model function, $\bm f(y)$ is obtained using Eq. (\ref{EM1}) and acceleration measurements are obtained using Eq. (\ref{eq:acc}). Since for measurement, accelerations of all DOF are considered, measurement model for the UKF remains same as acceleration model and can be written as,
\begin{equation}
\bm {h(y)} = \left[\begin{matrix}-\frac{1}{m_1}\left({y_{1}\,\left(k_{1}+k_{2}\right)-c_{2}\,y_{4}-k_{2}\,y_{3}+y_{2}\,\left(c_{1}+c_{2}\right)}\right)\\
\frac{1}{m_2}\left({c_{2}\,y_{2}-y_{3}\,\left(k_{2}+k_{3}\right)+c_{3}\,y_{6}+k_{2}\,y_{1}+k_{3}\,y_{5}-y_{4}\,\left(c_{2}+c_{3}\right)}\right)\\
-\frac{1}{m_3}\left({k_{4}\,y_{7}-c_{4}\,y_{8}-k_{3}\,y_{3}-c_{3}\,y_{4}+y_{5}\,\left(k_{3}-k_{4}\right)+\alpha_{DVP} \,{\left(y_{5}-y_{7}\right)}^3+y_{6}\,\left(c_{3}+c_{4}\right)}\right)\\ \frac{1}{m_4}\left({c_{4}\,y_{6}+c_{5}\,y_{10}-k_{4}\,y_{5}+k_{5}\,y_{9}+y_{7}\,\left(k_{4}-k_{5}\right)+\alpha_{DVP} \,{\left(y_{5}-y_{7}\right)}^3-y_{8}\,\left(c_{4}+c_{5}\right)}\right)\\ \frac{1}{m_5}\left({c_{5}\,y_{8}-y_{9}\,\left(k_{5}+k_{6}\right)+c_{6}\,y_{12}+k_{5}\,y_{7}+k_{6}\,y_{11}-y_{10}\,\left(c_{5}+c_{6}\right)}\right)\\ \frac{1}{m_6}\left({c_{6}\,y_{10}-y_{11}\,\left(k_{6}+k_{7}\right)+c_{7}\,y_{14}+k_{6}\,y_{9}+k_{7}\,y_{13}-y_{12}\,\left(c_{6}+c_{7}\right)}\right)\\ \frac{1}{m_7}\left({c_{7}\,y_{12}-c_{7}\,y_{14}+k_{7}\,y_{11}-k_{7}\,y_{13}}\right)\end{matrix}\right]
\end{equation}
Process noise co-variance matrix $\mathbf{Q}$ is obtained using the same process as discussed for 2-DOF system (refer Eq. (\ref{pnc:Q})) and is written as,
\begin{equation}
\begin{array}{c}
    \bm {q_c} = \sqrt{dt}\,\, diag\Large\left[\begin{smallmatrix}0& \frac{\sigma_1}{m_1}& 0& \frac{\sigma_2}{m_2}& 0& \frac{\sigma_3}{m_3}& 0& \frac{m_k^-(7)\,\,\sigma_4}{m_4}& 0& \frac{\sigma_5}{m_5}& 0& \frac{\sigma_6}{m_6}& 0& \frac{\sigma_7}{m_7}& 0& 0& 0& 0& 0& 0& 0\end{smallmatrix}\right]\\
    \mathbf{Q} = q_cq_c^T
\end{array}
\end{equation}
\noindent Where, $m_k^-(7)$ is the seventh element of UKF's predicted mean calculated from Eq. (\ref{pm-ukf}). The acceleration measurements and applied force are corrupted with a Gaussian noise having SNR values of 50 and 20 respectively. A comparison of the acceleration response obtained from data simulation and used in filtering is presented in Fig. \ref{acc-f-7dof}.

\begin{figure}[ht!]
    \centering
    \includegraphics[scale = 0.4]{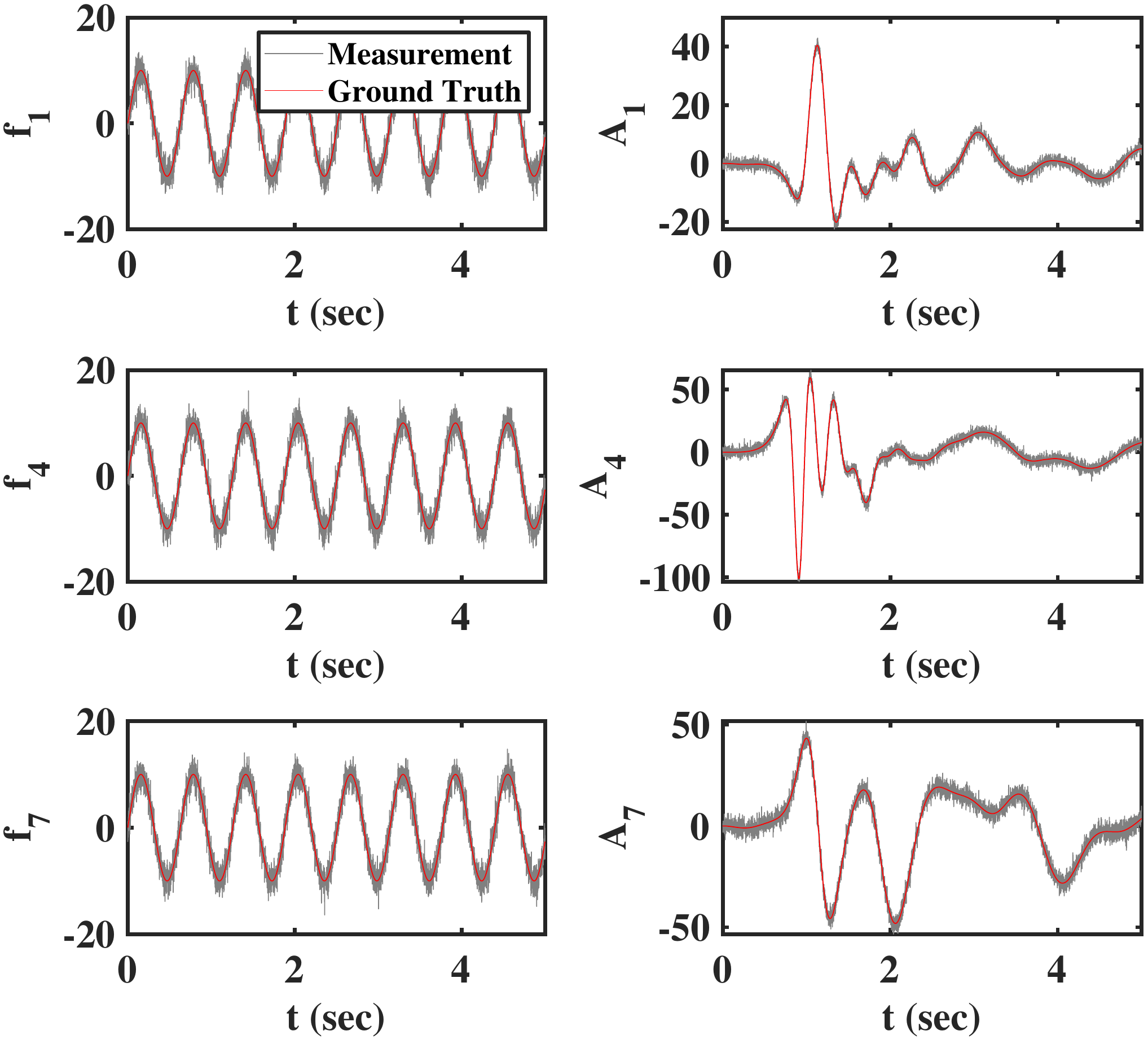}
    \caption{Deterministic component of force and acceleration vector corresponding to DOF 1, 4 and 7 used in UKF. The noisy acceleration vectors are provided as measurements to the UKF algorithm.}
    \label{fa-7dof-1dp-1}
\end{figure}

\begin{figure}[ht!]
\begin{subfigure}{0.45\textwidth}
    \centering
    \includegraphics[scale = 0.35]{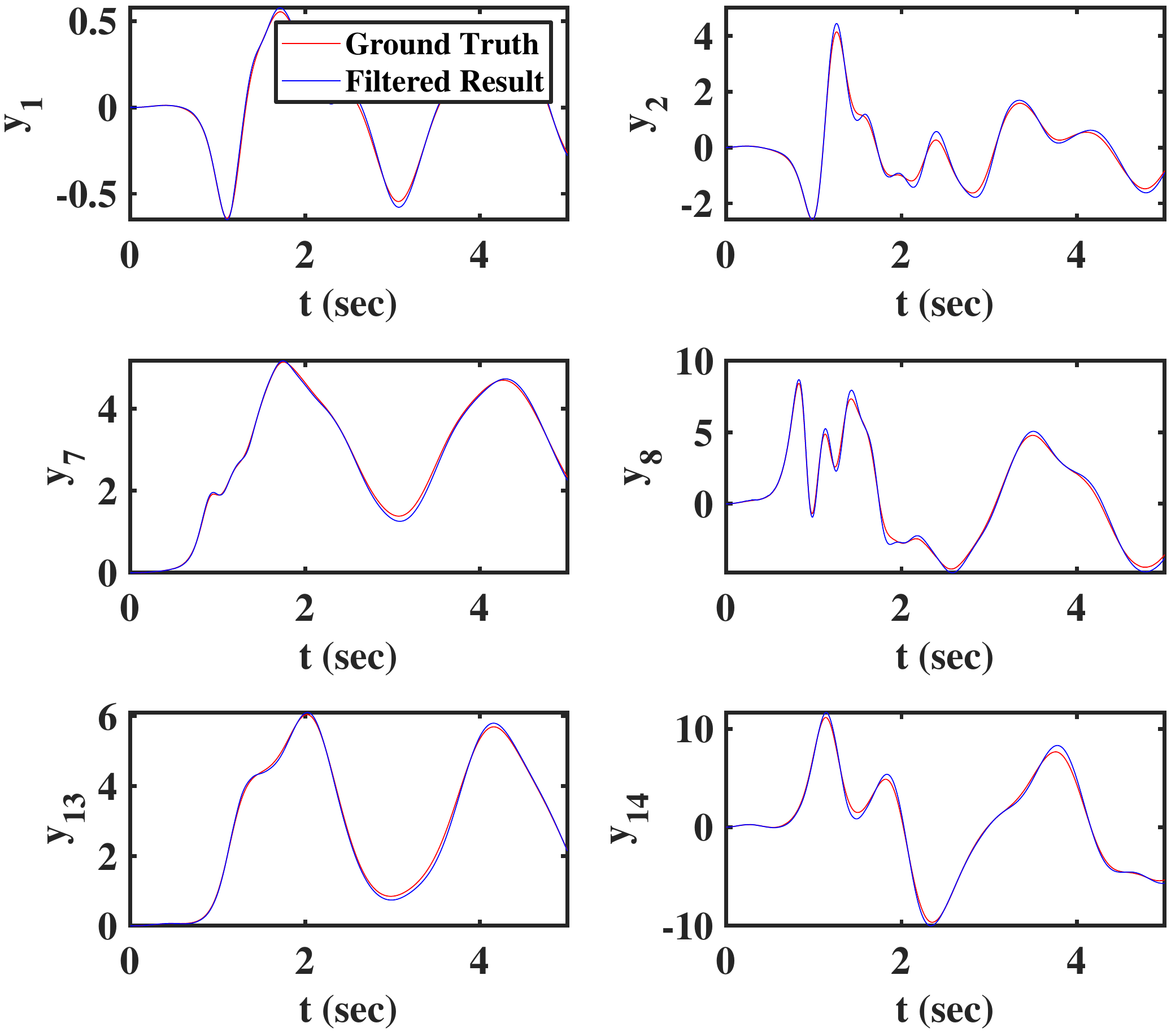}
    \caption{State (Displacement And Velocity) Estimation}
\end{subfigure}
\begin{subfigure}{0.45\textwidth}
    \centering
    \includegraphics[scale = 0.35]{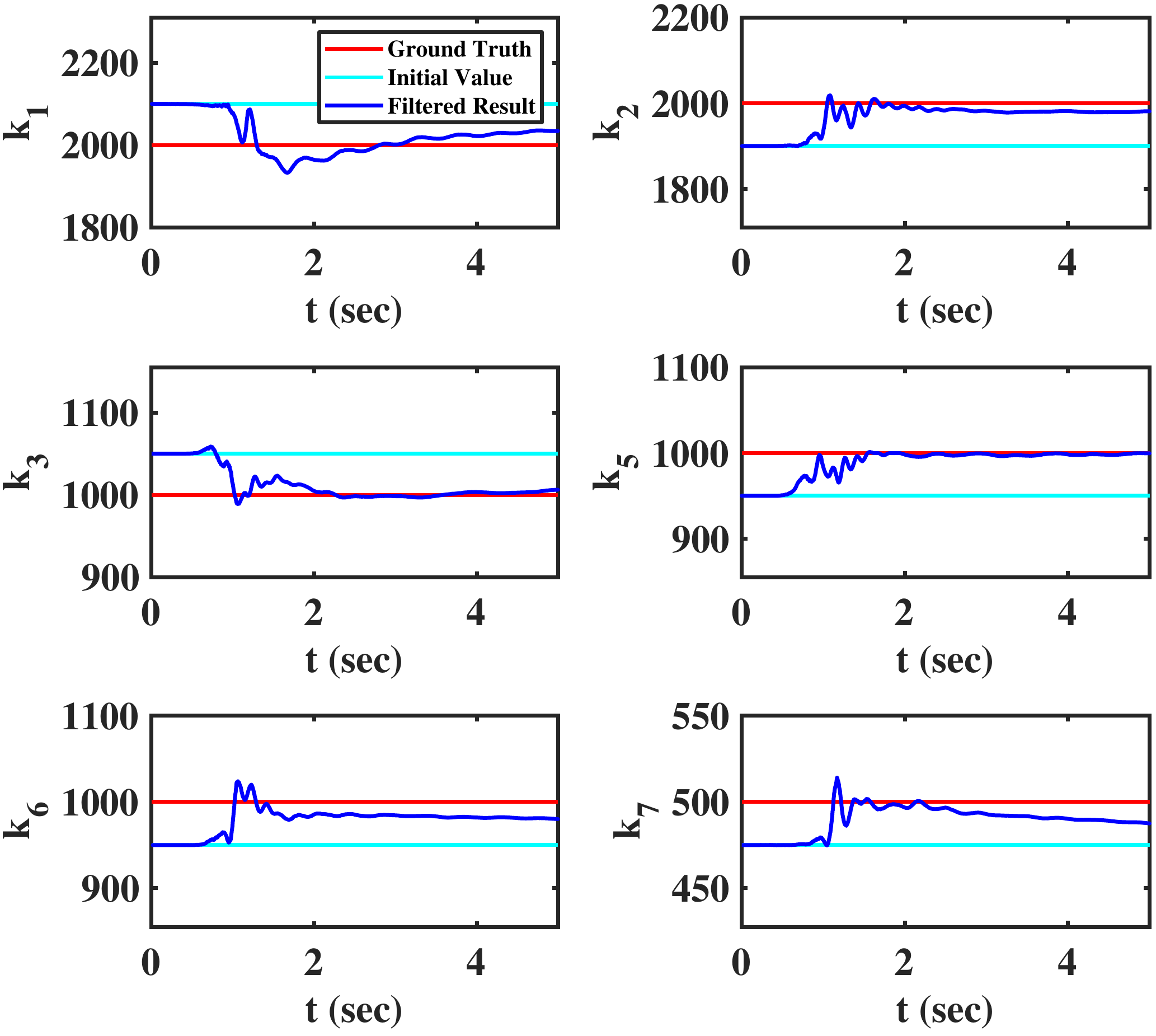}
    \caption{Parameter (Stiffness) Estimation}
\end{subfigure}
\caption{Combined state and parameter estimation results for the 7-DOF van der pol system.}
\label{7dof-1dp-1}
\end{figure}

\begin{figure}[ht!]
\begin{subfigure}{0.45\textwidth}
    \centering
    \includegraphics[scale = 0.25]{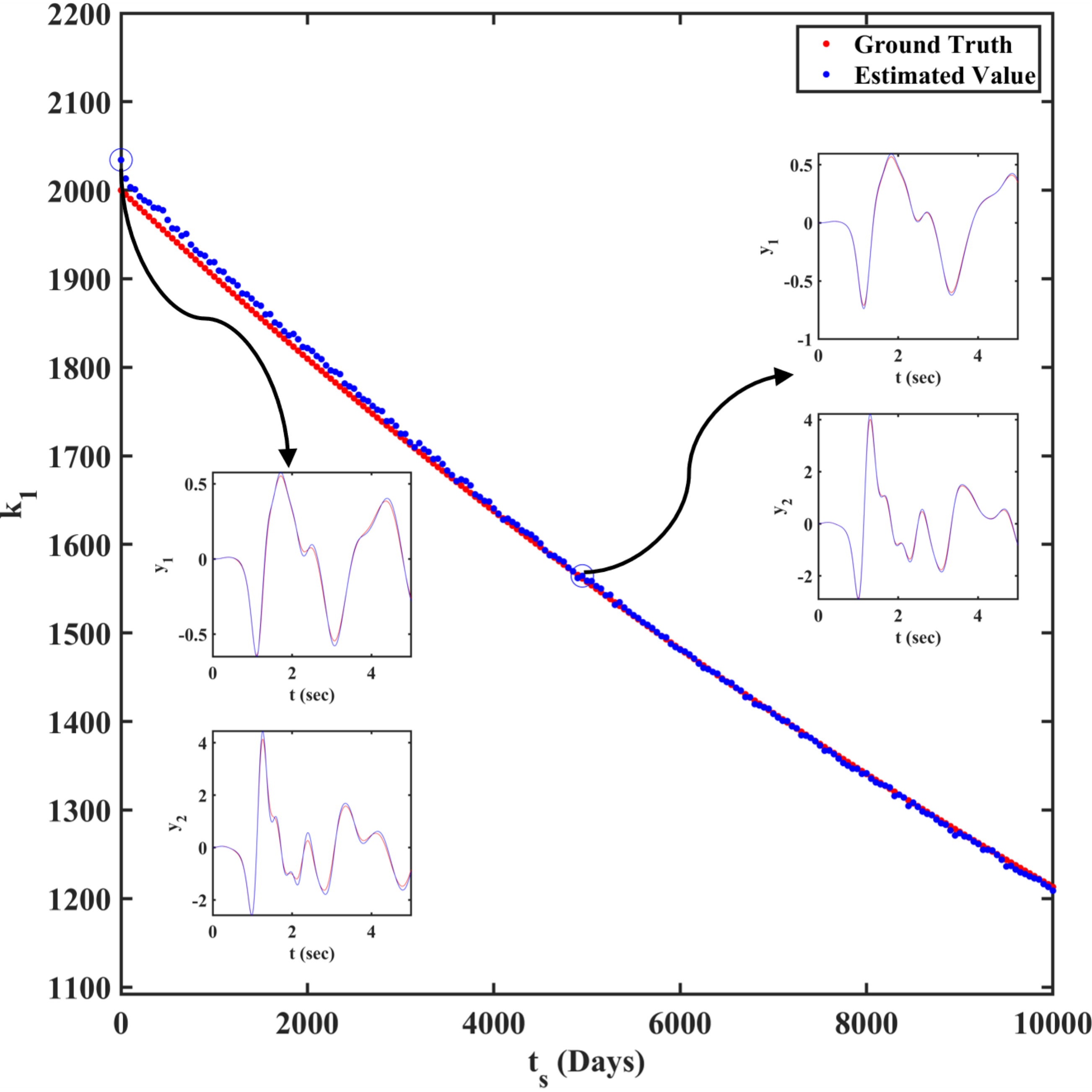}
    \caption{k1}
\end{subfigure}
\begin{subfigure}{0.45\textwidth}
    \centering
    \includegraphics[scale = 0.25]{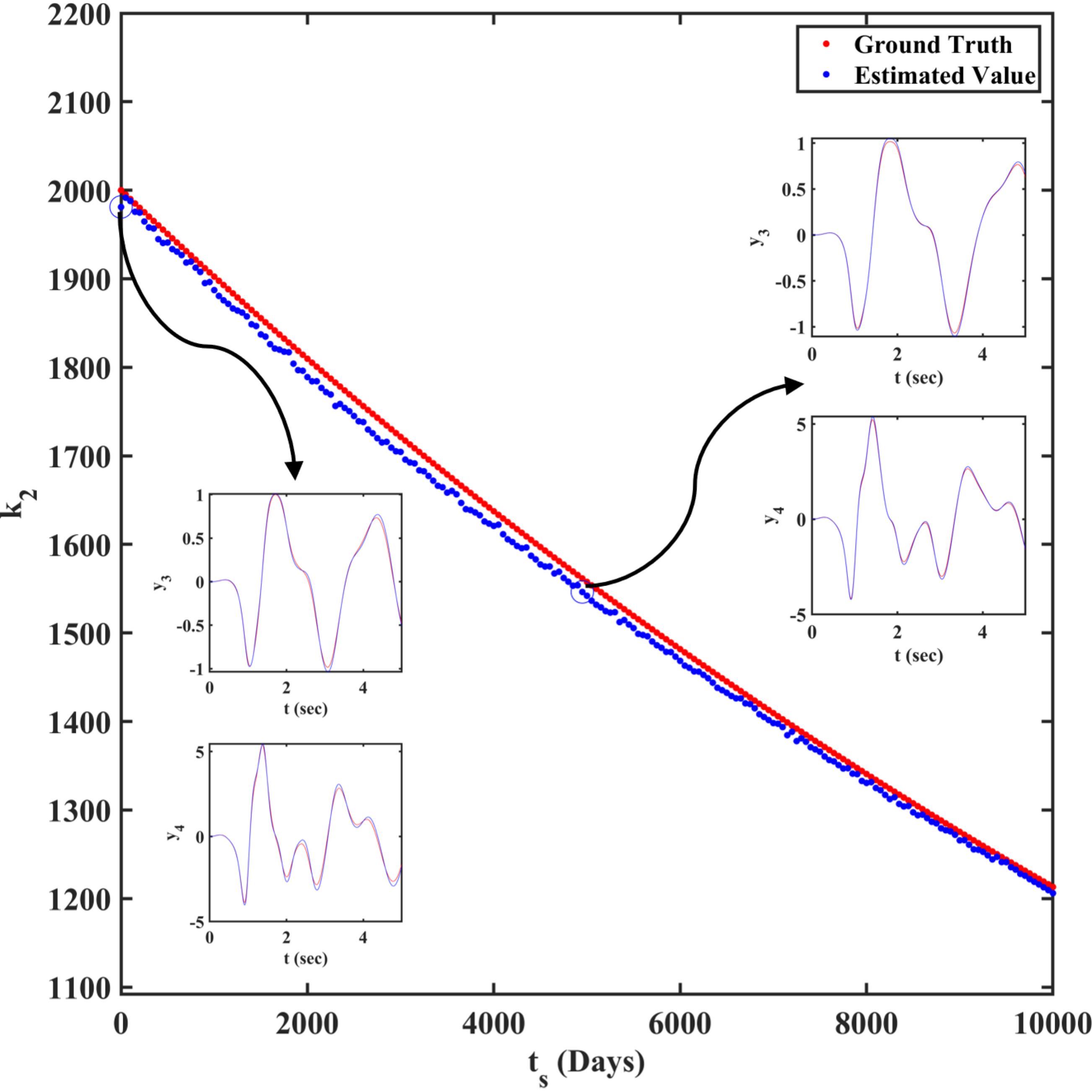}
    \caption{k2}
\end{subfigure}
\caption{Estimated stiffness ($k_1$ and $k_2$) in slow-time-scale using the UKF algorithm for the 7DOF example. State estimations at selected time-steps are also shown. Good match between the ground truth and the filtered result is obtained. These data act as input to the Gaussian process (GP).}
\label{f7-1}
\end{figure}

\begin{figure}[ht!]
\begin{subfigure}{1\textwidth}
    \centering
    \includegraphics[scale = 0.35]{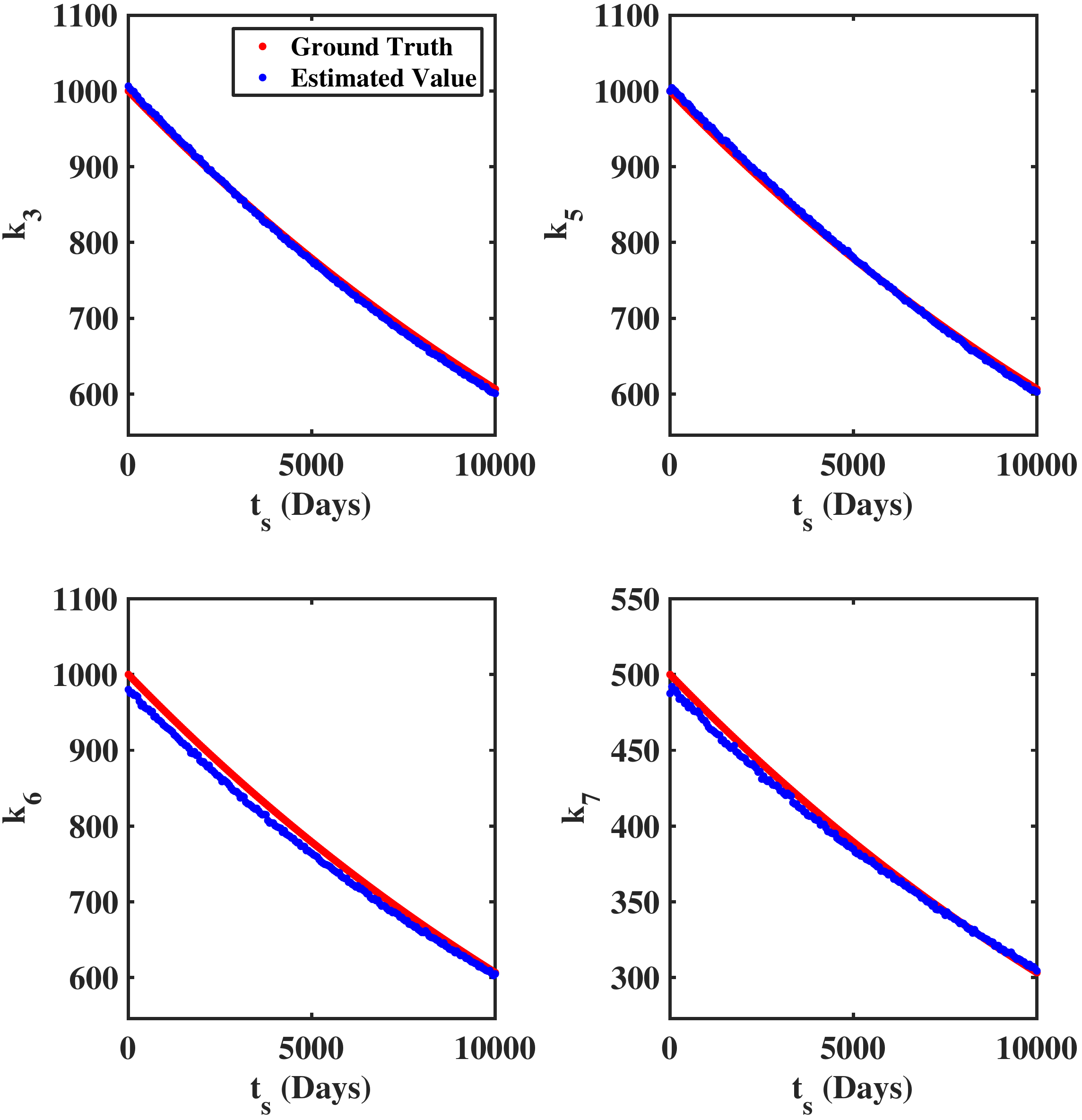}
\end{subfigure}
\caption{Estimated stiffness ($k_3$, $k_5$, $k_6$ and $k_7$) in slow-time-scale using the UKF algorithm for the 7DOF example. Good match between the ground truth and the filtered result is obtained. These data act as input to the Gaussian process (GP).}
\label{f7-2}
\end{figure}

\begin{figure}[ht!]
    \centering
    \includegraphics[scale = 0.35]{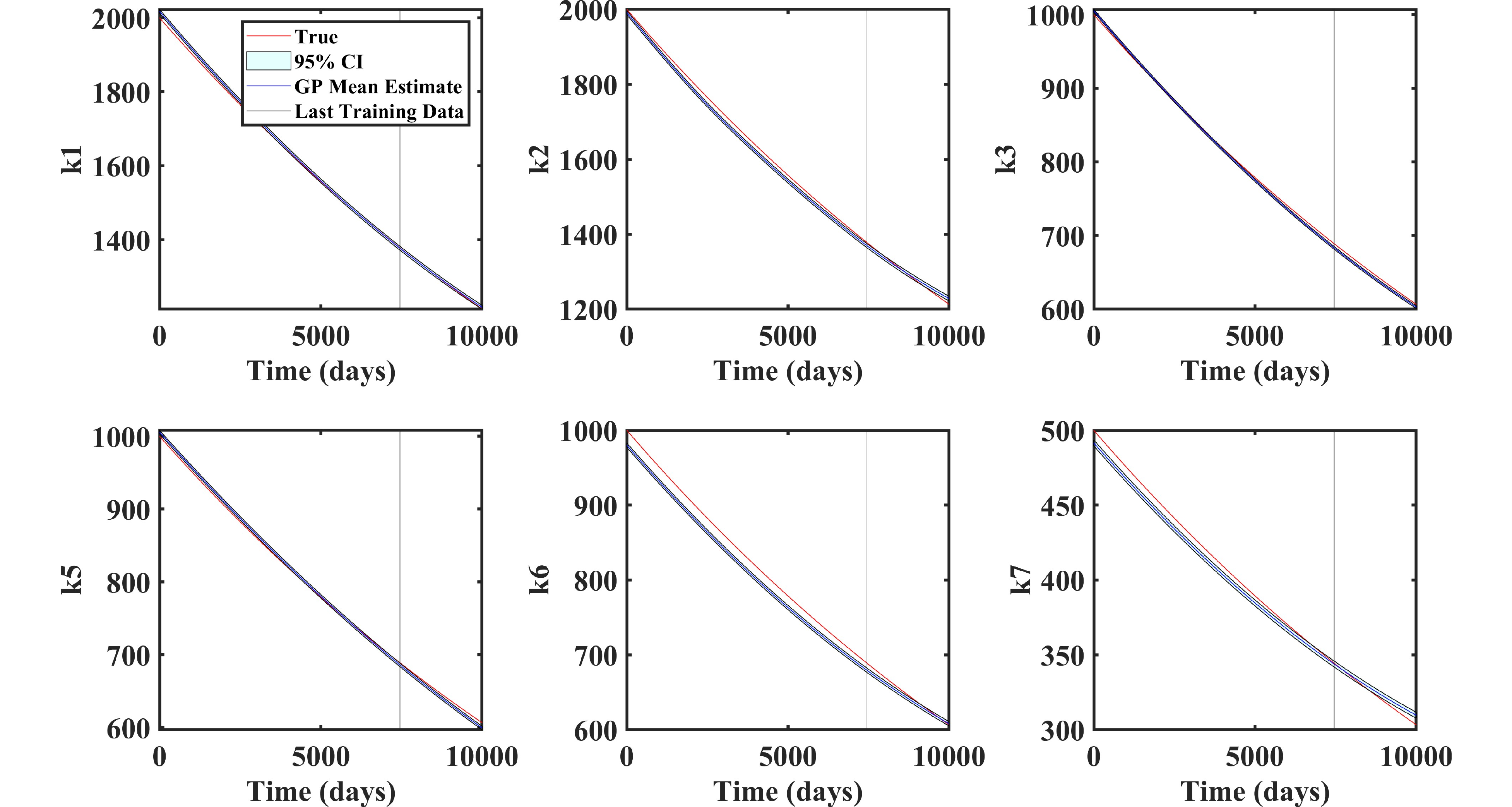}
    \caption{Results representing the performance of the proposed digital twin for the 7DOF system.  The GP is trained using the data generated using UKF. Data upto the horizontal line is available to the GP. The digital twin performs well even when predicting system parameters at future time-steps.}
    \label{gp7}
\end{figure}

Similar to the previous example, we first examine the performance of the UKF algorithm.
To that end, the acceleration vectors (noisy) shown in Fig. \ref{fa-7dof-1dp-1} is considered as the measurements.
The state and parameter estimation results obtained using the UKF algorithm are shown in Fig. \ref{7dof-1dp-1}. 
It can be observed that the proposed approach yields highly accurate estimate of the state vectors.
As for the parameter estimation, $k_2$, $k_3$ and $k_5$ converge exactly towards their respective true values.
As for  $k_1$, $k_6$ and $k_7$, UKF yields an accuracy of around 95\%.
A summary of the estimated parameters in the slow time-scales is shown in Fig. \ref{f7-1} and Fig. \ref{f7-2}. We observe that the estimates for new data points improve as our initial guess of system parameter improves (which for our case is the final parameters obtained from previous data points). Similar to Fig. \ref{7dof-1dp-1}, we observe that the estimates for stiffness $k_2$, $k_3$ and $k_5$ are more accurate than those obtained for $k_1$, $k_6$ and $k_7$.
These data is used for training the GP model.
Fig. \ref{gp7} shows the results obtained using the GP. 
The vertical line in Fig. \ref{gp7} indicate the point until which data is available to the GP.
For $k_1, k_2, k_3$ and $k_5$, the results obtained using GP matches exactly with the true solution.
For $k_7$, the GP predicted results are found to diverge from the true solution. However, the divergence is observed approximately after 3.5 years from the last observation, which for all practical purpose is sufficient for condition based maintenance.
For stiffness $k_6$ also, even though the filter estimates are less accurate at earlier time steps, the predicted results manage to give a good estimates of actual value which goes to show that if digital twin is given a regular stream of data, it has the capacity for self correction which in-turn helps better representation of the physical systems.  
\section{Conclusions}\label{conclusion}
The potential of digital twin in dynamical systems is immense; it can be used for health-monitoring, diagnosis, prognosis, active control and remaining useful life computation.
However, practical adaptation of this technology has been slower than expected, particularly because of insufficient application-specific details.
To address this issue, we propose a novel digital twin framework for stochastic nonlinear multi degree of freedom dynamical systems.
The proposed digital twin has four components -- 
(a) a physics-based nominal model (low-fidelity), (b) a Bayesian filtering algorithm a (c) a supervised machine learning algorithm
and (d) a high-fidelity model for predicting future responses.
The physics-based nominal model combined with Bayesian filtering is used for combined parameter-state estimation, and the GP is used for learning the temporal evolution of the parameters.
While the proposed framework can be used with any choice of Bayesian filtering and machine learning algorithm, the proposed approach uses unscented Kalman filter and Gaussian process in this paper.

Applicability of the proposed digital twin is illustrated with two stochastic nonlinear MDOF systems.
For both examples, we have assumed availability of acceleration measurements and the stochasticity is present in the applied force. 
In order to simulate a realistic scenario, a high-fidelity model (Taylor 1.5 strong) is used for data generation and a low-fidelity model (Euler Maruyama) is used for filtering.
The synthetic measurement data generated are corrupted with white Gaussian noise.
Cases pertaining to partial measurements (measurement at only selected degrees of freedom) and complete measurement (measurements at all degrees of freedom) are shown.
For all the cases, the proposed digital twin is found to yield highly accurate results with accuracy of 95\% and above, indicating its possible application to other realistic systems.

\section*{Acknowledgements}
AG and BH gratefully acknowledges the financial support received from Science and Engineering Research Board (SERB), Department of Science and Technology (DST), Government of India, (under the project no. IMP/2019/000276). SC acknowledges the financial support received from I-Hub foundation for Cobotics (IHFC) through seed funding.

\end{document}